\documentclass{article}

\PassOptionsToPackage{numbers, sort&compress}{natbib}

 \usepackage[main, final]{neurips_2026}
\usepackage[utf8]{inputenc} % allow utf-8 input
\usepackage[T1]{fontenc}    % use 8-bit T1 fonts
\usepackage{amsfonts}       % blackboard math symbols
\usepackage{microtype}
\usepackage{graphicx}
\usepackage{subfigure}

\usepackage{hyperref}       % hyperlinks
\usepackage{url}            % simple URL typesetting
\usepackage{nicefrac} 

\usepackage{tabularx}
\usepackage{dsfont}
\usepackage{multirow}
\usepackage{wrapfig}

\usepackage{multirow}
\usepackage{color}
\usepackage{lipsum,booktabs}
\usepackage[dvipsnames,table,xcdraw]{xcolor}

\usepackage{multicol}

\usepackage{caption}
\usepackage{subcaption}

\usepackage{enumitem}

\usepackage{upgreek}

\usepackage{array}
\usepackage{wrapfig}
\newcolumntype{C}[1]{>{\centering\arraybackslash}p{#1}}

\usepackage{amsmath,amsfonts,bm}

\def\eqref#1{equation~\ref{#1}}
\def\1{\bm{1}}

\DeclareMathAlphabet{\mathsfit}{\encodingdefault}{\sfdefault}{m}{sl}
\SetMathAlphabet{\mathsfit}{bold}{\encodingdefault}{\sfdefault}{bx}{n}

\DeclareMathOperator*{\argmax}{arg\,max}

\usepackage{amsmath}
\usepackage{amssymb}
\usepackage{mathtools}
\usepackage{amsthm}
\usepackage{bbding}

\usepackage{makecell}  % for better cell formatting
\usepackage{adjustbox} % for fixed-width rotated box

\definecolor{brightmaroon}{rgb}{0.76, 0.13, 0.28}
\definecolor{darkcyan}{rgb}{0.0, 0.55, 0.55}

\definecolor{best}{HTML}{1F77B4}    % blue
\definecolor{second}{HTML}{FF0000}  % gray

\usepackage[capitalize,noabbrev]{cleveref}

\theoremstyle{plain}

\theoremstyle{definition}

\theoremstyle{remark}

\usepackage[textsize=tiny]{todonotes}

\title{\textit{Golden Layers and Where to Find Them}:\\ Improved Knowledge Editing for Large Language Models Via Layer Gradient Analysis}

\usepackage{authblk}

\author[1]{Shrestha Datta}
\author[2]{Hongfu Liu}
\author[1]{Anshuman Chhabra\thanks{Corresponding Author.}\hspace{0.35em}}

\affil[1]{University of South Florida, Tampa, FL, USA}
\affil[2]{Brandeis University, Waltham, MA, USA}
\affil[ ]{\texttt{\{shresthadatta, anshumanc\}@usf.edu, hongfuliu@brandeis.edu}}

\begin{document}

\maketitle

\begin{abstract}
    Knowledge editing in Large Language Models (LLMs) aims to update the model’s prediction for a specific query to a desired target while preserving its behavior on all other inputs. This process typically involves two stages: identifying the layer to edit and performing the parameter update. Intuitively, different queries may localize knowledge at different depths of the model, resulting in different sample-wise editing performance for a fixed editing layer. In this work, we hypothesize the existence of fixed \textit{golden} layers that can achieve \textit{near-optimal} editing performance similar to sample-wise optimal layers. To validate this hypothesis, we provide empirical evidence by comparing golden layers against ground-truth sample-wise optimal layers. Furthermore, we show that golden layers can be reliably identified using a proxy dataset and generalize effectively to unseen test set queries across datasets. Finally, we propose a novel method, namely \textit{Layer Gradient Analysis (LGA)}, that estimates golden layers efficiently via gradient-attribution, avoiding extensive trial-and-error across multiple editing runs. Extensive experiments on several benchmark datasets demonstrate the effectiveness and robustness of our LGA approach across different LLM types and various knowledge editing methods.  
\end{abstract}

\section{Introduction}
\vspace{-0.75mm}
\looseness-1Large Language Models (LLMs) encode extensive factual and relational knowledge acquired during pre-training, enabling strong performance across a wide range of downstream tasks. However, this knowledge is implicitly distributed across a large number of model parameters, making it expensive to directly access or update after the pre-training stage. As a result, LLMs may produce outdated, incorrect, or undesirable outputs, and rectifying such errors through full retraining or large-scale fine-tuning is often computationally prohibitive. Moreover, doing so can also potentially degrade previously learned desirable behaviors and useful model knowledge unrelated to the information that needs to be edited \cite{kirkpatrick2017overcoming, luo2025empirical}. Knowledge editing methods \cite{yao2023editinglargelanguagemodels} in LLMs thus seek to address this challenge by modifying a model’s prediction for a specific test query to a desired target while preserving its behavior on all other inputs. This capability is increasingly important for real-world deployment, where models must be adaptable, reliable, and aligned with evolving real-world information.

\looseness-1Most existing knowledge editing approaches follow a \textit{locate-then-edit} paradigm \cite{wang2024easyediteasytouseknowledgeediting}. Given a test query and a desired new target knowledge, knowledge editing methods first identify where the relevant knowledge is stored in the model and then perform a constrained parameter update at that location. In practice, this typically reduces to selecting a \textit{layer} (or in some cases, block of layers) to edit, followed by an optimization-based update to those parameters. Prior work has shown that the choice of editing layer is a critical factor in determining both rewrite success and the degree of unintended effect to other unrelated model knowledge \cite{takahashi2025understanding, hase2023does}. {Consequently, several methods attempt to identify important layers using causal signals, such as \textit{Causal Mediation Analysis (CMA)} \cite{meng2022locatingRome}, which has emerged as a widely adopted standard; or via gradients, such as Salient Layers Editing Model (SaLEM) \cite{mishra-etal-2024-correcting}.} Despite the perceived effectiveness of these methods, prior work has shown that these strategies often fail to reliably identify the best editing layer \cite{hase2023does, hase2024interpretable}, motivating the need for improved layer-wise editing performance prediction methods.   

\looseness-1 Therefore, in this work, we take a closer look at how editing performance varies across layers and uncover a consistent empirical pattern. While the sample-wise optimal editing layer may differ across individual queries, we observe that, at the dataset level, a large majority of samples concentrate their optimal performance on the same layer or a small set of layers. This observation motivates us to define the concept of \textit{golden layers}, which we define as fixed layers that, when used uniformly across samples, achieve aggregate editing performance that is statistically indistinguishable and/or near-optimal to that obtained by editing each sample at its own optimal layer. The existence of golden layers suggests that knowledge relevant for editing is not evenly distributed across all model layers, and that certain layers act as particularly effective intervention points. Importantly, this phenomenon implies that near-optimal editing performance can be achieved without per-sample layer selection, provided that these layers can be identified. 

% At the same time, existing layer selection methods such as CMA are not designed to detect such globally optimal layers, leading to both unnecessary computation and a reduction in performance.

Further, motivated by these observations, in this paper, we propose \textit{Layer Gradient Analysis (LGA)}, a novel and efficient method for estimating golden layers without performing any actual knowledge edits. Inspired by recent work on sample-wise gradient attribution \cite{jiao2025feasibility, chhabra2025oga, pruthi2020tracin}, LGA leverages \textit{layer-specific} gradient attribution to quantify how strongly each layer mediates the interaction between the model’s existing knowledge and the desired new target knowledge. By aggregating first-order gradient signals across a proxy set of editing queries, LGA produces a robust estimate of which layers are most suitable for knowledge editing. This approach avoids exhaustive layer-wise trial-and-error, scales efficiently to large models, and directly targets the layers that yield near-optimal editing performance in practice. Through extensive experiments across multiple LLM architectures, datasets, and editing methods, we show that editing at layers selected by LGA consistently outperforms standard layer selection approaches such as CMA and SaLEM, while significantly reducing the computational overhead associated with them.

In sum, we highlight our major contributions in this work:\vspace{-1mm}
\begin{itemize}[nosep, leftmargin=2em]
    \looseness-1\item We empirically demonstrate the existence of \textit{golden layers} for knowledge editing in LLMs, showing that in most cases, fixed editing layers can achieve near-optimal or statistically indistinguishable performance compared to sample-wise optimal layer selection across datasets and models. \vspace{1mm}
    \item Furthermore, we show that golden layers can be reliably estimated using proxy datasets and generalize effectively to unseen test queries, enabling potential editing strategies that are more computationally efficient than exhaustive layer-wise trial-and-error. \vspace{1mm}%and performant than traditional methods such as Causal Mediation Analysis (CMA).
    \item To this end, we propose a novel first-order, gradient-attribution-based method for identifying golden layers, named Layer Gradient Analysis (LGA). Through extensive experiments across LLMs, editing strategies, and benchmark datasets, we demonstrate that LGA outperforms CMA and SaLEM in terms of both editing performance as well as computational efficiency.
\end{itemize}

\vspace{-0.75mm}

\section{Related Work} \vspace{-0.75mm}
\textbf{Locate-Then-Edit Knowledge Editing.} \looseness-1 There are several popular locate-then-edit methods that have been developed for knowledge editing in LLMs.  \textit{ROME} \cite{meng2022locatingRome} performs editing by applying a rank-one update to the weights of a selected MLP layer, inserting a new key-value association that encodes the desired factual modification while approximately preserving previously stored associations. Building on \textit{ROME}, \textit{R-ROME} \cite{gupta2024rebuilding} addresses instability in sequential editing by enforcing consistent use of averaged key representations in the rank-one update formulation. \textit{MEMIT} \cite{meng2023memit} extends single-fact editing methods to the mass-editing regime by simultaneously inserting thousands of factual associations via a batched least-squares update to transformer weights. \textit{EMMET} \cite{gupta2024emmet} further generalizes this framework by formulating batched editing under equality constraints, providing a closed-form solution that unifies \textit{ROME} and \textit{MEMIT} through a preservation-memorization objective. Adjacent to this line of work, \textit{PMET} \cite{li2024pmet} adopts an optimization-based approach that jointly optimizes hidden states in both the attention and feedforward modules. It is important to note that across all these methods, the selection of editing layers is typically guided by CMA \cite{meng2022locatingRome}, which we will discuss next.

\looseness-1\textbf{Limitations of Causal Mediation Analysis.} A common strategy in locate-then-edit knowledge editing methods is to select MLP intervention layers using CMA (sometimes also referred to as \textit{causal tracing}), under the assumption that components identified as mediating factual recall will be effective targets for editing. CMA treats the transformer network as a causal graph over hidden activations and quantifies how intermediate components mediate the effect of an input query on the model’s prediction \cite{pearl2022direct, vig2020investigating}. However, recent work provides a detailed empirical critique of this assumption, showing that CMA layer scores are often poorly correlated with editing success across a variety of knowledge editing problem variants, and that this weak correlation persists across multiple problem settings \cite{hase2023does}. Follow-up work \cite{hase2024interpretable} further corroborates this across additional models and evaluation setups, demonstrating that the choice of \textit{editing layer}, rather than causal tracing scores, remains a stronger predictor of editing success. Complementary work~\cite{nishi2024representationshattering} examines the mechanistic underpinnings of editing interventions, showing that while CMA attributes decisive roles to early-to-mid MLP sublayers in factual recall, the impact of an edit on the learned representation manifold (e.g., distortion or shattering of entity subspaces) is not well predicted by these attribution scores. These findings highlight the limitations of CMA for layer selection in LLMs and motivate our work on developing efficient layer-selection strategies that are more robust and performant for editing success.

\looseness-1\textbf{Gradient-Based Attribution.} While gradient-based attribution methods such as saliency maps \cite{adebayo2018sanity} and integrated gradients \cite{sundararajan2017axiomatic} have had long-standing success in interpretability research, recent work on data valuation has demonstrated their potential as effective estimators of training sample impact on model predictions \cite{koh2017understanding, yang2024revisit, chhabra2024what, pruthi2020tracin}. More specifically, first-order methods only utilize sample-gradients (obtainable in one-pass) to undertake this analysis \cite{bejan2023make, chhabra2025oga, pruthi2020tracin}, balancing both performance and computational efficiency. Recent work has also sought to utilize gradients to assess layer impact in applications such as pruning, mixture-of-experts allocation \cite{askari2025layerif}, influence analysis \cite{vitel2025first, yeh2022first}, among others. Moreover, for knowledge editing, SaLEM \cite{mishra-etal-2024-correcting} also utilizes first-order information for layer selection, but as our results will subsequently show, it fails to utilize the full power of the gradient signal. Thus, in our paper, we bridge this gap by proposing the novel and efficient first-order Layer Gradient Analysis (LGA) method as an alternative to CMA-based layer selection for model-agnostic editing. 

\looseness-1\textbf{Memorization.} Identifying fixed layers for KE is closely related to memorization. Prior work finds that memorization is not spread evenly across a network: certain layers are impacted by it rather than all layers \cite{stephenson2021geometry}, it is concentrated in a small set of neurons scattered over multiple layers \cite{maini2023memorization}, and, in LLMs, it can be localized to specific weights that, when fine-tuned, make the model forget the memorized data \cite{stoehr2024localizing}. These findings indicate the possibility of existence of multiple layers for a successful edit in KE. Besides, Memorization in LLMs can be localized by comparing the gradients of memorized and non-memorized paragraphs \cite{stoehr2024localizing}. This motivates our use of both old- and new-knowledge gradients for layer selection.
\vspace{-0.75mm}
\section{Research Questions} \vspace{-0.5mm}
\looseness-1 Knowledge editing aims to modify a model $M$ such that for a specific query $Q$ prediction/knowledge ($\hat{K} = M(Q)$) is updated to a desired target $K_{\text{new}}$, while preserving the model’s behavior on all other inputs. That is, after applying a knowledge editing process $\mathcal{E}$ on the original model $M$, we wish to obtain an edited model $M'$ which satisfies $M'(Q) = K_{\text{new}}$ while ensuring $M(Q') = M'(Q')$, for all queries $Q' \neq Q$. It typically consists of two phases: (i) identifying the layer to edit, and (ii) modifying the parameters of the selected layer. After selecting a layer block $L$, the editing procedure $\mathcal{E}$ produces a knowledge-edited model $M' = \mathcal{E}(M; L, Q, K_{\text{new}})$. Different editing methods formalize \( \mathcal{E} \) using various optimization-based or closed-form methods \cite{meng2022locatingRome, gupta2024rebuilding, meng2023memit, gupta2024emmet, li2024pmet}, but they share the common objective of enforcing the target behavior specified by $(Q, K_{\text{new}})$ while limiting collateral changes outside the edited layers. 

Moreover, as demonstrated in past work \cite{hase2023does}, the choice of editing layer plays a significant role in final editing performance. Specifically, in our investigations on knowledge localization, we observe an intriguing phenomenon: despite the diversity of the underlying information or knowledge, a majority of samples consistently select the same editing layer. This observation motivates us to explore the following two fundamental questions:
\begin{itemize}[nosep,leftmargin=2em]
\looseness-1\item \textit{Does there exist a fixed “golden” layer, or a small set of fixed “golden” layers, that achieves near-optimal performance for knowledge editing when compared to the sample-wise optimal layer?} \vspace{0.2mm}
\item \textit{If such a golden layer exists, can it be efficiently identified without resorting to extensive trial-and-error across multiple editing runs?} 
\end{itemize}

% \vspace{-1.2mm}

\looseness-1 The identification of such a golden layer is of immense importance to the knowledge editing community and, more broadly, to research on LLMs. Moreover, existing methods for layer selection have been shown to fail at detecting the optimal editing layer \cite{hase2023does, hase2024interpretable}. Therefore, they have to depend on sample-wise or layer-wise trial-and-error to select a potential layer for editing, leading to substantial computational overhead and limited scalability. A fixed or near-universal golden layer would eliminate repeated layer searches, enabling more efficient, stable, and reproducible knowledge edits. Beyond efficiency, golden layer identification also offers theoretical value. While prior work \cite{meng2022locatingRome, geva2021transformer, geva2023dissecting, geva2022transformer} largely assumes that knowledge in LLMs is stored as key–value representations distributed across layers, recent studies \cite{men2025shortgpt, ju2024large, wei2024does, he2024matters} have begun to question this view, suggesting that knowledge may be redundantly or unevenly encoded. By investigating the existence of golden layers, our work provides new empirical evidence on how knowledge is organized across model layers, thereby offering a principled foundation for more effective and interpretable knowledge editing methods.

Motivated by these questions, we structure our paper into two parts. First, we empirically investigate the existence of golden layers through extensive trial-and-error experiments. Second, we propose a novel gradient-attribution-based algorithm to estimate the impact of each layer on knowledge editing, enabling efficient golden-layer identification while avoiding costly repeated computations.

% \begin{table*}[t]
% \centering
% \small
% %\renewcommand{\arraystretch}{0.95}
% \setlength{\tabcolsep}{1.5pt}
% \caption{The mean performance of the Optimal Layer (top performing layer for a test set sample) and Golden Layer (selected using the proxy set), for each of the editing performance metrics on the \textit{ZSRE} dataset. The mean of the gaps between the performance of the Optimal Layer and Golden Layer is also reported. Hypothesis testing undertaken via the Kolmogorov-Smirnov test found that the Optimal and Golden Layer average performance distributions were not statistically significant for all metrics.}
% \begin{tabular}{l c c c | c c c | c c c}
% \toprule
% Metric & \multicolumn{3}{c}{GPT-2 XL} & \multicolumn{3}{c}{LLaMA2-7B} & \multicolumn{3}{c}{Gemma3-12B}\\
% \cline{2-10}
%  & Optimal & Golden & Diff. & Optimal & Golden & Diff. & Optimal & Golden & Diff.\\
% \midrule
% Rewrite     & 1.0000 & 1.0000 & 0.0070 & 0.9691 & 0.9667 & 0.0024 & 0.9997 & 0.9763 & 0.0235 \\
% Rephrase    & 0.9749 & 0.8838 & 0.0910 & 0.9678 & 0.9176 & 0.0511 & 0.9983 & 0.9294 & 0.0689 \\
% Locality    & 1.0000 & 0.9959 & 0.0041 & 0.9999 & 0.9927 & 0.0072 & 0.9995 & 0.9795 & 0.0200 \\
% Portability & 0.6022 & 0.4844 & 0.1178 & 0.7326 & 0.5993 & 0.1332 & 0.7094 & 0.5480 & 0.1614 \\\bottomrule
% Overall     & 0.8524 & 0.7361 & 0.1164 & 0.9614 & 0.8725 & 0.0889 & 0.9545 & 0.8426 & 0.1119 \\
% \bottomrule
% \end{tabular}
% \label{tab:oracle_vs_golden}
% \end{table*}

\vspace{-1.5mm}
\section{The Existence of Golden Layers} \vspace{-1mm}
In this section, we detail the process of golden layer discovery. We first formalize the definition of a golden layer and provide empirical evidence of its existence by comparing it with the ground-truth, sample-wise optimal editing layer. We then demonstrate that the golden layer for a target dataset can be reliably estimated using a related proxy dataset (or even an independent dataset), showing that while the golden layer is largely invariant to individual samples, it systematically varies across different LLM architectures.

Throughout the paper, we conduct experiments across three different LLMs: GPT-2 XL \cite{radford2019language}, LLaMA2-7B \cite{touvron2023llama}, and Gemma3-12B \cite{team2025gemma}, and evaluate on several commonly used knowledge editing benchmark datasets: \textit{ZSRE} \cite{meng2022locatingRome, levy2017zero}, \textit{WikiBio} \cite{hartvigsen2023aging}, \textit{WikiCounterfact} \cite{zhang2024comprehensive}, \textit{WikiRecent} \cite{cohen2024evaluating}, \textit{Counterfact} \cite{meng2022locatingRome}. Moreover, we evaluate editing performance using standard metrics: \textit{Rewrite Accuracy} ($\uparrow$), \textit{Rephrase Accuracy} ($\uparrow$), \textit{Locality} ($\uparrow$), \textit{Portability} ($\uparrow$), and \textit{Fluency} ($\uparrow$), but note that not all metrics are supported on all datasets. Additionally, for easy overall comparison we also provide an \textit{Overall} metric that is simply a weighted average of all other normalized metrics, where higher values denote better performance.\footnote{Appendices~\ref{app:implementation}, \ref{app:code}, and \ref{app:metrics} provide details regarding implementation, experimental setup, and the performance metrics of the editing pipeline.}

\subsection{Defining Golden Layers} 
Following prior literature \cite{meng2022locatingRome, gupta2024rebuilding, gupta2024emmet} and practical considerations, knowledge editing is typically performed on a single layer, with the goal of incorporating new knowledge while preserving existing and broadly shared knowledge. Given access to ground truth, the sample-wise optimal editing layer can be obtained by exhaustively performing knowledge edits at each layer and selecting the one that yields the best editing performance. At the individual sample level, different samples may exhibit different optimal editing layers. However, at the group level, namely, over a dataset with a sufficient number of samples, we observe that a majority of samples consistently favor the same editing layer. Motivated by this empirical regularity, we define golden layers as follows:

\textbf{Definition \textit{(Golden Layers)}.} \textit{Given an LLM and a test set of samples for knowledge editing, \textit{golden layers} are fixed layers for editing across all those samples that achieve, in aggregate, similar performance from that obtained by editing each sample at its own sample-wise optimal layer.}

Based on the above definition, we formulate our first research question: \textit{do golden layers exist?} To examine this hypothesis, we conduct experiments on several benchmark datasets \cite{levy2017zero, wang2024easyediteasytouseknowledgeediting}. Specifically, we exhaustively search for the sample-wise optimal editing layer for each individual query instance, according to \textit{Rewrite Accuracy}, the most important metric in knowledge editing, and in parallel identify a single fixed layer that achieves the best average performance across the entire benchmark. By comparing the performance of this fixed layer with that of the sample-wise optimal layers, we empirically evaluate whether a golden layer can approximate the optimal editing behavior at the dataset level.

\begin{table*}[t]
\centering
\vspace{4mm}
\small
{
\renewcommand{\arraystretch}{1} % increase row height globall
\setlength{\tabcolsep}{1pt}
\caption{\looseness-1 The performance of the Sample-Wise Optimal Layer (top performing layer for a sample in the test set) and Golden Layer (optimal fixed layer for all samples derived from the test set), for \textit{ZSRE}, \textit{WikiBio}, \textit{WikiCounterfact}, \textit{WikiRecent}, and \textit{Counterfact} datasets with R-ROME on three LLMs: GPT-2 XL, LLaMA2-7B, and Gemma3-12B. The optimal layer selection is based on \textit{Rewrite Accuracy.} The two performances are found to be statistically different using the two-sided Student's t-test.}\label{tab:oracle_vs_golden_datasets}
\vspace{-1mm}
\resizebox{\linewidth}{!}{

\begin{tabular}{c| c |C{2cm} C{2cm} C{2cm} C{2cm} C{2cm}}
\toprule
Model & Method & \textit{ZSRE} & \textit{WikiBio} & \textit{WikiCounterfact} & \textit{WikiRecent} & \textit{Counterfact}\\
\midrule
\multirow{3}{*}{\rotatebox{90}{\small GPT-2}}&Sample-Wise Optimal Layer & 1.0000$_{\pm0.0000}$ & 0.8897$_{\pm0.1239}$ & 0.9860$_{\pm0.0721}$ & 0.9974$_{\pm0.0425}$ & 1.0000$_{\pm0.0000}$\\
        &Golden Layer & 0.9993$_{\pm0.0135}$ & 0.8256$_{\pm0.1629}$ & 0.9537$_{\pm0.1480}$ & 0.9887$_{\pm0.0793}$ & 0.9871$_{\pm0.1128}$\\
        \cmidrule{2-7}
&Statistical Difference & No & Yes & Yes & Yes & Yes \\
\midrule
\multirow{3}{*}{\rotatebox{90}{LLaMA2}}&Sample-Wise Optimal Layer & 0.9691$_{\pm0.0747}$ & 1.0000$_{\pm0.0006}$ & 0.9949$_{\pm0.0345}$ & 0.9897$_{\pm0.0677}$ & 1.0000$_{\pm0.0000}$\\
        &Golden Layer & 0.9667$_{\pm0.0771}$ & 0.9912$_{\pm0.0325}$ & 0.9922$_{\pm0.0532}$ & 0.9792$_{\pm0.0910}$ & 0.9995$_{\pm0.0164}$\\
        \cmidrule{2-7}
&Statistical Difference & No & Yes & No & Yes & No \\
\midrule
\multirow{3}{*}{\rotatebox{90}{Gemma3}}&Sample-Wise Optimal Layer & 0.9997$_{\pm0.0082}$ & 0.9296$_{\pm0.0827}$ & 0.9443$_{\pm0.1333}$ & 0.9991$_{\pm0.0143}$ & 1.0000$_{\pm0.0000}$\\
        &Golden Layer & 0.9763$_{\pm0.0908}$ & 0.8538$_{\pm0.1208}$ & 0.9415$_{\pm0.1625}$ & 0.9789$_{\pm0.0950}$ & 1.0000$_{\pm0.0000}$\\
        \cmidrule{2-7}
&Statistical Difference & Yes & Yes & No & Yes & No \\

\bottomrule
\end{tabular}
}
}
\end{table*}

\begin{figure*}[t]
    \centering
    \vspace{4mm}
    \includegraphics[width=0.99\textwidth]{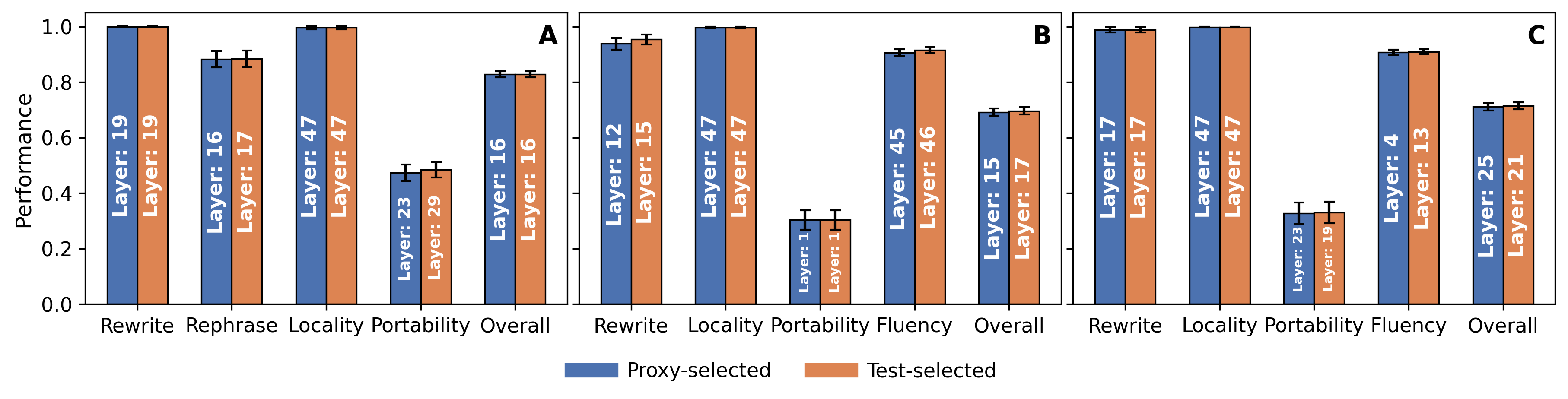}\vspace{-3mm}
    \caption{Performance of golden layers selected via the proxy and test sets with GPT-2 XL on (A) \textit{ZSRE}, (B) \textit{WikiCounterfact}, and (C) {\textit{Counterfact}}. Editing performance evaluation is conducted on the test set queries. Error bars denote the standard error of the mean.}\vspace{-1mm}
    \label{fig:proxy_vs_test}
\end{figure*}

 % (results for other metrics are provided in Appendix~\ref{app:sample-optimal-golden-others} due to space limitations)
\looseness-1 Table~\ref{tab:oracle_vs_golden_datasets} compares the knowledge editing performance of the sample-wise optimal layer and the golden layer across all benchmark datasets and LLMs, using \textit{Rewrite Accuracy} as the evaluation metric. For fair comparison, both settings employ the R-ROME editing method~\cite{gupta2024rebuilding}.
According to the Student’s t-test, no statistically significant difference is observed between the golden layer and the sample-wise optimal layer in five cases, including GPT-2 XL on \textit{ZSRE}, LLaMA2-7B on \textit{ZSRE} and \textit{WikiCounterfact}, and Gemma3-12B on \textit{WikiCounterfact} and \textit{Counterfact}. Although the statistical test does not pass in the remaining cases, the absolute performance gaps are generally small with the golden layers mostly achieving a \textit{Rewrite Accuracy} exceeding 0.95, which is typically sufficient for practical deployment. Notable exceptions occur on \textit{WikiBio} with GPT-2 XL and Gemma3-12B, where the golden layer exhibits lower performance with 0.8256 and 0.8538 relative to the sample-wise optimal values. Nevertheless, even in these cases, the performance remains competitive: a strong baseline, CMA~\cite{meng2022locatingRome}, achieves \textit{Rewrite Accuracies} of only 0.6894 and 0.6817, respectively. These results suggest that a fixed golden layer can serve as a reasonable approximation to sample-wise layer selection for knowledge editing, while substantially reducing the computational overhead associated with exhaustive per-sample layer searches.

\vspace{-1.75mm}
\subsection{Golden Layers Across Datasets}
\looseness-1Above, we have demonstrated the existence of golden layers on the test set. However, in practical settings, the test set is typically unavailable during training or deployment. To address this limitation, we investigate whether golden layers can be identified using a proxy dataset as a surrogate. Specifically, we aim to estimate the golden layer on an accessible proxy set and examine whether it can effectively generalize to unseen test sets. To achieve this, we first seek a related proxy dataset and exhaustively check the layer performance in terms of knowledge editing, and select the layer with the best performance as the golden layer. In subsequent sections, we will propose an efficient method for golden layer estimation, obviating such expensive trial-and-error analysis.%Note that we will propose an efficient algorithm to estimate the golden layer in next section to avoid the extensive trial-and-error computing.  

\looseness-1 Figure~\ref{fig:proxy_vs_test} presents the knowledge editing performance of GPT-2 XL when using golden layers identified on a proxy set versus a test set across three benchmark datasets: \textit{ZSRE}, \textit{WikiCounterfact}, and {\textit{Counterfact}}.\footnote{Due to space constraints, we provide additional results for \textit{WikiBio} and \textit{{WikiRecent}} on GPT-2 XL in Appendix~\ref{app:proxy_vs_test} and for all datasets on LLaMA2-7B and Gemma3-12B in Appendix~\ref{app:proxy-vs-test-others}.} In these experiments, the proxy and test sets are randomly split from the same dataset (10\%: proxy; 90\%: test). We observe that the golden layers identified from the proxy set and the test set often coincide. {For example, on \textit{ZSRE} and \textit{Counterfact}, the golden layers selected for the \textit{Overall} and \textit{Rewrite} metric, respectively, are identical across the two splits.} Even in cases where the selected golden layers differ, the resulting editing performance remains very close. These results indicate that golden layers identified from an accessible proxy set can effectively generalize to unseen test sets. Furthermore, the observations suggest that multiple layers may exhibit comparable near-optimal performance, implying the existence of more than one potential golden layer. 

\begin{figure*}[t]
    \centering
    \includegraphics[width=1\textwidth]{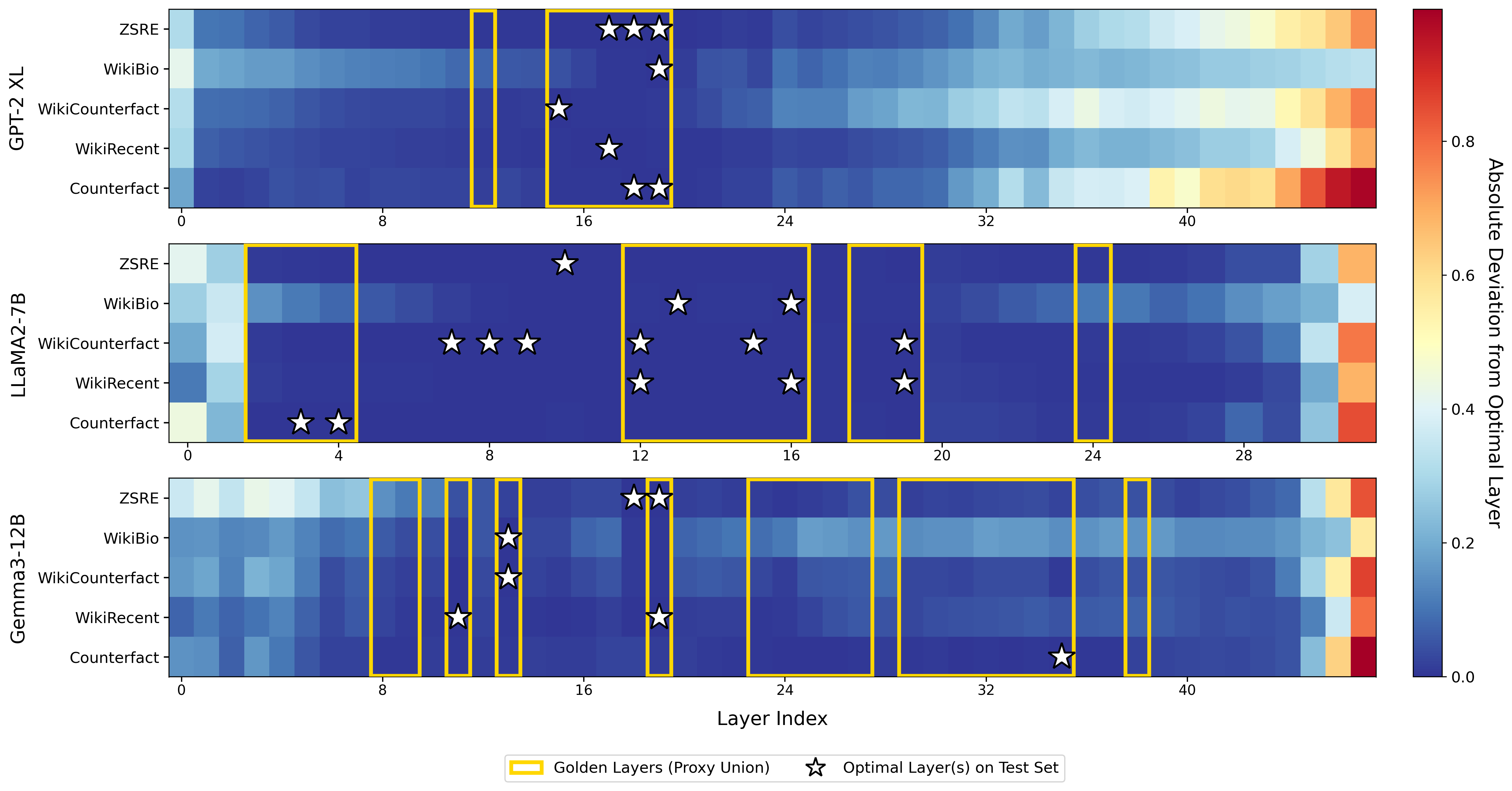}\vspace{-2mm}
    \caption{Visualization of model layers for GPT-2 XL, LLaMA2-7B, and Gemma3-12B, where each cell indicates the data-specific performance of a layer measured as the absolute deviation in \textit{Rewrite Accuracy} performance from the optimal layer on the test set. Darker blue cells indicate better performing layers with {\FiveStarOpen} denoting optimal layers (these can be tied in performance with multiple optimal layers on the same dataset). The golden cells across layers denote the union of the golden layers independently identified using the proxy set of each dataset. This union of proxy-set golden layers generally select the higher performing layers, and most often, the optimal layers themselves. %Reduce font size and remove bold and reduce cell height -- best layer on test set 
    % Heatmap of absolute deviation from the optimal test-selected layer for \textbf{Rewrite Accuracy}.
    % Blue indicates layers close to the test-set optimum, while red indicates larger deviation.
    % Yellow outlines denote the union of proxy-selected optimal performing layers across datasets (golden region),
    % green outlines indicate dataset-specific proxy-selected best performing layers, and the star marks the optimal
    % test-selected layer(s).
    }
    \label{fig:rewrite_heatmap_proxy_vs_test}\vspace{-3mm}
\end{figure*}

\looseness-1To further examine the robustness of golden layers identified using a proxy set, we construct a proxy union that aggregates five proxy-estimated golden layer subsets, each drawn from the corresponding benchmark datasets. We then evaluate whether they generalize to the corresponding individual test sets. Figure~\ref{fig:rewrite_heatmap_proxy_vs_test} presents a heatmap of \textit{Rewrite Accuracy}, comparing the golden layers identified from the proxy union with the optimal layers on each test set. Darker blue regions indicate smaller performance gaps, whereas red regions correspond to larger discrepancies. Stars denote the optimal layers on the test sets; note that multiple layers may be optimal when they achieve identical performance. The yellow bands highlight the golden layers estimated from the proxy union. As shown in the figure, the golden layers derived from the proxy union largely overlap with the test-set optimal layers, particularly for GPT-2 XL and LLaMA2-7B. This alignment suggests that golden layers estimated from an aggregated proxy set can generalize reasonably well to unseen test sets, supporting their robustness across datasets. Moreover, even though the proxy layers are drawn from different datasets, they can still identify the optimal layer or layers with \textit{Rewrite Accuracy} close \footnote{To further corroborate the cross-dataset generalizability of the proxy-estimated golden layers, we provide additional results in Appendix~\ref{app:lodo} owing to space constraints.} to the optimal layer. Interestingly, as Figure \ref{fig:rewrite_heatmap_proxy_vs_test} demonstrates, while golden layers overlap across optimal test-set layers for the same LLM, they appear in different parts of the network for different LLMs. This implies that golden layers are a \textit{model-dependent} property, while being largely \textit{data-independent}.

\section{Golden Layer Estimation Using Layer Gradient Analysis (LGA)}

In the previous sections, we have verified the existence of golden layers, and their potential benefits in serving as the editing layers for knowledge editing. Subsequently, efficient and accurate estimation of golden layers then becomes a key next step towards unlocking their potential in editing pipelines. Towards this goal, we now propose a novel golden layer estimation method that efficiently only operates on first-order information (i.e., layer-specific gradients), and identifies highly performant golden layers. Through the layers selected using our \textit{Layer Gradient Analysis (LGA)} method, we consistently attain higher editing performance compared to standard methods for knowledge editing, such as \textit{Causal Mediation Analysis (CMA)} \cite{meng2022locatingRome} and {\textit{Salient Layers Editing Model (SaLEM)} \cite{mishra-etal-2024-correcting}. %We next describe LGA and its inner workings, and then present our findings demonstrating exceptional editing performance and computational efficiency of LGA over CMA.

\subsection{The Proposed Layer Gradient Analysis Approach} \label{sec:lga_approach}

%have to propose and define the gradient-based layer attribution methods that you will be using

Our proposed approach for accurately estimating golden is based on analyzing layer-specific gradients, motivated by recent work on gradient-based data attribution \cite{chhabra2025oga, yang2024revisit, pruthi2020tracin}, which essentially utilize sample gradients to assess how a training sample $z$ impacts the performance of a model (parameterized by weights $\hat{\theta}$ trained using loss $l$) on a validation/test sample $v$, simply using their inner product \cite{pruthi2020tracin, yang2024revisit}, defined as: $\phi(z,v)$$=$$\nabla_{\hat{\theta}} \ell(\hat{\theta}; z)^\top$$\cdot$$\nabla_{\hat{\theta}} \ell(\hat{\theta}; v)$. Higher gradient similarity values denote more training sample \textit{impact} on validation/test sample performance. Also note that this process is computationally efficient as gradients can be obtained in one-pass from the model post-training.

\looseness-1In this paper, we extend the gradient attribution inner product from the sample-level to the layer-specific impact. The key idea lies in realizing that the inner product can be restricted to specific layer weights, thereby enabling an assessment of how that specific layer block impacts model performance. That is, $\phi_{{L}}(z,v)$$=$$\nabla_{\theta_{{L}}} \ell(\hat{\theta}; z)^\top$$\cdot$$
\nabla_{\theta_{{L}}} \ell(\hat{\theta}; v)$, where $\nabla_{\theta_{{L}}}\ell(\hat{\theta}; z)$ denotes the sample $z$'s gradient of the $L$-th layer. Moreover, by aggregating these layer-specific attribution scores for a set of samples, we can obtain robust estimates for the layer's impact on the downstream task as $\sum_{(z, v)\in Z}\phi_{{L}}(z,v)$. These scores can then be used to compare between layers in terms of how beneficial they are for a task. This framework forms the basis of our Layer Gradient Analysis (LGA) approach.

Clearly, the loss function and the choice of samples will play a significant role in defining the task for which we wish to undertake the layer-attribution analysis. We now propose the procedure for golden layer estimation for knowledge editing in LLMs. For instance, consider, the LLM parameterized by network weights $\hat{\theta}$ trained using autoregressive cross entropy loss $\ell$. Moreover, consider a set of editing queries in the proxy set $\mathcal{Q} = \{Q_i\}_{i=1}^n$ and the old model knowledge associated with query $Q_i$ as $K_i$ and the new target knowledge as $K'_i$. Intuitively, we posit that utilizing the gradients related to both the new and old knowledge for a query can help identify the impact of the layer in the knowledge editing task. To this end, we can estimate the impact of layers across all proxy set queries and predict the golden layer $G^*$ through LGA:\vspace{-6mm}

\begin{align*}\label{eq:lga}
    G^* &= \argmax_{L \in \mathcal{L}}\sum_{Q_i \in \mathcal{Q}}\phi_{{L}}(Q_i \cup K_i,Q_i \cup K'_i),\\
    &= \argmax_{L \in \mathcal{L}}\sum_{Q_i \in \mathcal{Q}}\nabla_{\theta_{{L}}} \ell(\hat{\theta}; Q_i \cup K_i)^\top\cdot
\nabla_{\theta_{{L}}} \ell(\hat{\theta}; Q_i \cup K'_i).\vspace{-3mm}
\end{align*}
We provide the intuition behind each term of the LGA criterion and its formal justification in Appendix~\ref{app:lga_justification}. Now that we have obtained $G^*$ using our LGA method, we can utilize a knowledge editing method $\mathcal{E}$ to undertake the editing process on a test query $(Q_t, K'_t)$ as $\mathcal{E}(M; G^*, Q_t, K'_t)$. In the subsequent sections we present our findings for estimating golden layers and how that leads to superior editing performance compared to traditional layer identification methods, such as CMA.

\subsection{Golden Layer Identification and Knowledge Editing Performance Comparison}

We now compare editing performance when editing layer selection is conducted using LGA, CMA, and SaLEM. Note that our experimental setup is the same as in the previous section. \vspace{1mm}

\begin{table*}[t]  
\centering
\vspace{4mm}
\caption{\looseness-1 Editing performance evaluation of LGA, SaLEM, and CMA across different knowledge editing methods and the \textit{ZSRE}, \textit{WikiBio}, \textit{WikiCounterFact}, \textit{Counterfact} datasets on GPT-2 XL with R-ROME editing method. Different datasets support different performance metrics, where \textit{Rewrite Accuracy} $\rightarrow$ RwA,  \textit{Rephrase Accuracy} $\rightarrow$ RpA, \textit{Locality} $\rightarrow$ LOC, \textit{Portability} $\rightarrow$ PRT, \textit{Fluency} $\rightarrow$ FLC, and \textit{Overall} $\rightarrow$ OV. Layers are identified using the proxy set for both methods and evaluation is undertaken on an unseen test set, demonstrating performance improvements attained by LGA over baselines.}
\label{tab:combined_tables_editing_methods}\vspace{-1mm}
\footnotesize
\begin{minipage}{0.5\textwidth}
\centering
\textit{{ZSRE}}
\medskip
{
\setlength{\tabcolsep}{1.04pt}
\resizebox{\linewidth}{!}{
\begin{tabular}{c | c | c c c c c}
\toprule
Edit & Selection & RwA & RpA & LOC & PRT & OV \\
\midrule

\multirow{3}{*}{\shortstack[c]{R-ROME}}
& CMA & 0.9665 & 0.7467 & {0.9608} & {0.4837} &  0.6942 \\
& SaLEM & {0.6910} & {0.5951} & 0.6405 & {0.3629} & 0.5724 \\
& LGA (Ours) & {0.9851} & {0.8151} & 0.9543 & {0.4757} & \textbf{0.7106} \\
\midrule
\multirow{3}{*}{EMMET}
& CMA & {0.9850} & {0.8796} & {0.6589} & {0.4202} & {0.7359} \\
& SaLEM & 0.8072&  0.5816&   0.7042&      0.4441&  0.6343\\
& LGA (Ours) & {0.9862} & {0.8833} & {0.6857} & {0.3947} &  \textbf{0.7375} \\
\midrule
\multirow{3}{*}{ROME}
& CMA & 0.9669 & 0.7464 & 0.9604 & 0.4832  & 0.7892 \\
& SaLEM & 0.6776 & 0.5765 & 0.6302 & 0.3633  & 0.5619 \\
& LGA (Ours) & 0.9862 & 0.8178 & 0.9559 & 0.4778  & \textbf{0.8094} \\

\bottomrule
\end{tabular}
}
}
\end{minipage}%
\hfill
\begin{minipage}{0.5\textwidth}  % right column
\centering
{\textit{WikiBio}}
\medskip
{
\setlength{\tabcolsep}{3.3pt}
\resizebox{\linewidth}{!}{
\begin{tabular}{c | c | c c c c}
\toprule
Edit & Selection & RwA & LOC & FLC & OV \\
\midrule

\multirow{3}{*}{R-ROME}
& CMA & {0.6894} & {0.5692} & 0.8927 & 0.7171 \\
& SaLEM & {0.4061} & 0.2563 & 0.8186 & 0.4937 \\
& LGA (Ours) & {0.7484} & 0.5587 & {0.8906} & \textbf{0.7326} \\
\midrule
\multirow{3}{*}{EMMET}
& CMA & 0.8258 & 0.3894 & 0.8612 & \textbf{0.6921} \\
& SaLEM & 0.5784 & 0.2811 & 0.8641 & 0.5745 \\
& LGA (Ours) & 0.8433  & 0.2793  & 0.8894 & 0.6707 \\
\midrule
\multirow{3}{*}{ROME}
& CMA & 0.6911  & 0.2902  & 0.8788 & 0.6200 \\
& SaLEM & 0.4123  & 0.1932  & 0.8214 & 0.4756 \\
& LGA (Ours) & 0.7487  & 0.2851  & 0.8840 & \textbf{0.6393} \\

\bottomrule
\end{tabular}
}
}
\end{minipage}

\vspace{-1mm}  % small vertical space between rows

\begin{minipage}{0.50\textwidth}  % bottom-left
\centering
{\textit{WikiCounterFact}}  % width=0, centered
\medskip
{
\setlength{\tabcolsep}{1.04pt}
\resizebox{\linewidth}{!}{
\begin{tabular}{c | c | c c c c c}
\toprule
Edit & Selection & RwA & LOC & PRT & FLC & OV \\
\midrule

\multirow{3}{*}{R-ROME}
& CMA & 0.9135 & 0.5601 & 0.2651 & 0.9080 & 0.6741 \\
& SaLEM & {0.6370} & 0.3198 & {0.2365} & 0.7629 & 0.4965 \\
& LGA (Ours) & {0.9330} & {0.5669} & {0.2663} & {0.9072} & \textbf{0.6808} \\
\midrule
\multirow{3}{*}{EMMET}
& CMA & 0.9713  & 0.3931 & 0.2695 & 0.8039 & 0.6201 \\
& SaLEM & 0.6962  & 0.4342 & 0.1813 & 0.8603 & 0.5538 \\
& LGA (Ours) & 0.9841  & 0.4075 & 0.2831 & 0.8343 & \textbf{0.6382} \\
\midrule
\multirow{3}{*}{ROME}
& CMA & 0.9150  & 0.5582 & 0.2635 & 0.9142 & 0.6749 \\
& SaLEM & 0.6299  & 0.3036 & 0.2281 & 0.7635 & 0.4891 \\
& LGA (Ours) & 0.9379  & 0.5695 & 0.2659 & 0.9118 & \textbf{0.6839} \\

\bottomrule
\end{tabular}
}
}
\end{minipage}%
\hfill
\begin{minipage}{0.5\textwidth}  % bottom-right
\centering
{{\textit{Counterfact}}}  % width=0, centered
\medskip
{
\setlength{\tabcolsep}{1.04pt}
\resizebox{\linewidth}{!}{
\begin{tabular}{c | c | c c c c c}
\toprule
Edit & Selection & RwA & RpA & LOC & PRT & OV \\
\midrule
\multirow{3}{*}{\shortstack[c]{R-ROME}}
& CMA &  {0.9420} &  {0.6284} & {0.9613} & {0.4387} &  {0.7426} \\
& SaLEM & 0.7981 & 0.5983 & 0.7111 & 0.3585 & 0.6165 \\
& LGA (Ours) & 0.9592 & 0.7154 & 0.9560 & 0.4362 & \textbf{0.7667} \\
\midrule
\multirow{3}{*}{EMMET}
& CMA & {0.9957} & {0.3330} & 0.5693 & {0.3905} & {0.5721} \\
& SaLEM & 0.9066 & 0.1359 & 0.6639 & 0.4153 & 0.5304 \\
& LGA (Ours) & 0.9925 & 0.3550 & {0.6565} & 0.4081 & \textbf{0.6030} \\
\midrule
\multirow{3}{*}{ROME}
& CMA & 0.9914 & 0.3545 & 0.8579 & 0.4409 & {0.6612} \\
& SaLEM & 0.7637 & 0.3131 & 0.5133 & 0.3568 & 0.4867 \\
& LGA (Ours) & 0.9968 & 0.3974 & {0.8667} & 0.4367 & \textbf{0.6744} \\

\bottomrule
\end{tabular}
}
}
\end{minipage}
\vspace{-6mm}
\end{table*}

\textbf{Comparing LGA, CMA, and SaLEM Across Different Editing Methods.}
\looseness-1 We first evaluate the consistency of our layer selection strategy LGA relative to CMA across different editing methods. We edit the GPT-2 XL model on the \textit{ZSRE}, \textit{WikiBio}, \textit{WikiCounterFact}, and \textit{Counterfact} datasets (results for \textit{WikiRecent} provided in Appendix~\ref{app:counterfact_methods_llms} due to space limitations). We employ three editing methods: R-ROME \cite{gupta2024rebuilding}, ROME \cite{meng2022locatingRome}, and EMMET \cite{gupta2024emmet} at the respective selected layers. The results are provided in Table~\ref{tab:combined_tables_editing_methods}. As can be observed from \textit{Overall} scores, layers identified by LGA yield higher performance than those selected by CMA and SaLEM across all the configurations, with the exception of EMMET being used with \textit{WikiBio}. %In general, when examining \textit{Overall} scores, LGA-selected layers consistently lead to improved performance for R-ROME across all datasets. A similar pattern is observed for ROME and EMMET, where LGA yields higher performance on \textit{ZSRE}, \textit{WikiCounterFact}, and \textit{WikiRecent}.
Our findings hence suggest that LGA provides a more stable layer selection strategy than CMA and SaLEM across editing methods and datasets.

\looseness-1 Across the editing methods, LGA consistently improves \textit{Rewrite Accuracy}, achieving a maximum improvement of 8.56\% and 84.29\% on \textit{WikiBio} compared to CMA and SaLEM, respectively. In addition, \textit{Rephrase Accuracy}, which is only applicable to \textit{ZSRE} and \textit{Counterfact}, also exhibits consistent gains under LGA regardless of the editing method, with the greatest improvement being 9.16\% and 161.22\%, respectively. In terms of aggregate performance trends across datasets, LGA exhibits improved \textit{Overall} score increases when applied with different editing methods, with 
% average gains of 2.13\% and 
average gains of 1.94\% and 29.40\% over CMA and SaLEM, respectively. Similar trends are observed across other metrics, such as \textit{Locality}, across datasets. In sum, LGA serves as a better layer selection mechanism compared to CMA and SaLEM, and attains improved performance across metrics and datasets. \vspace{1.5mm}

\textbf{LGA vs CMA vs SaLEM Performance Analysis Across LLMs.}
\looseness-1We now undertake comparative performance evaluation across different and diverse LLM architectures. We employ the R-ROME editing method owing to its superlative editing performance compared to other methods. We apply LGA, SaLEM, and CMA to GPT-2 XL, LLaMA2-7B, and Gemma3-12B on the \textit{ZSRE}, \textit{WikiBio}, \textit{WikiCounterFact}, and \textit{Counterfact} datasets using the R-ROME editing method (results for \textit{WikiRecent} provided in Appendix~\ref{app:counterfact_methods_llms} due to space constraints). We provide the results in Figure~\ref{fig:lga-cma}. As the figure demonstrates, in terms of \textit{Overall} performance, LGA consistently outperforms CMA across all datasets for GPT-2 XL. For LLaMA2-7B, LGA achieves higher \textit{Overall} performance on \textit{ZSRE}, \textit{WikiBio}, and \textit{Counterfact}, while attaining competitive performance on \textit{WikiCounterFact}. Furthermore, the gains are even more pronounced for Gemma3-12B, with significant \textit{Overall} performance improvement attained for \textit{ZSRE}, \textit{WikiBio}, and \textit{WikiCounterfact}. Similar trends hold for the comparison with SaLEM, as LGA attains improved performance on average. %\textcolor{red}{When comparing LGA to SaLEM, we observe that LGA slightly underperforms \textit{Overall} performance when editing LLaMA2-7B for \textit{ZSRE} \& \textit{WikiCounterfact} and Gemma3-12B for both \textit{ZSRE} \& \textit{WikiBio} instances with an average deficit of $1.095\%$), while in the remaining 8 scenarios LGA outperforms relative to SaLEM with an average gain of $18.785\%$.}

\begin{wraptable}{r}{0.55\linewidth}
% \vspace{-3mm}
\centering
\caption{\looseness-1 Editing performance comparison between LGA and random layer selection on \textit{ZSRE} with R-ROME measured by metrics: \textit{Rewrite Accuracy} $\rightarrow$ RwA,  \textit{Rephrase Accuracy} $\rightarrow$ RpA, \textit{Locality} $\rightarrow$ LOC, \textit{Portability} $\rightarrow$ PRT, and \textit{Overall} $\rightarrow$ OV. LGA achieves an average \textit{Overall} performance gain of 8.69\% across the three models.}
\label{tab:random_selection_comparison_main}\vspace{-1mm}
\footnotesize
{
\setlength{\tabcolsep}{1.04pt}
\resizebox{\linewidth}{!}{
\begin{tabular}{c | c | c c c c c}
\toprule
Model & Selection & RwA & RpA & LOC & PRT & OV \\
\midrule
\multirow{2}{*}{GPT-2 XL}
& Random & 0.8539 & 0.5916 & 0.9701 & 0.4784 & 0.6347 \\
& LGA (Ours) & 0.9851 & 0.8151 & 0.9543 & 0.4757 & \textbf{0.7106} \\
\midrule
\multirow{2}{*}{LLaMA2-7B}
& Random & 0.9576 & 0.8115 & 0.9885 & 0.5650 & 0.8307 \\
& LGA (Ours) & 0.9625 & 0.8988 & 0.9927 & 0.5950 & \textbf{0.8623} \\
\midrule
\multirow{2}{*}{Gemma3-12B}
& Random & 0.8197 & 0.7692 & 0.7988 & 0.4465 & 0.7086 \\
& LGA (Ours) & 0.9535 & 0.8796 & 0.7736 & 0.5198 & \textbf{0.7816} \\
\bottomrule
\end{tabular}
}
}
\vspace{-3mm}
\end{wraptable}

\looseness-1 With respect to \textit{Rewrite Accuracy}, LGA selected layers consistently yield higher performance compared to CMA, with improvements reaching up to 19.2\% on the \textit{WikiBio} dataset when editing the Gemma3-12B model. A similar pattern is observed for \textit{Rephrase Accuracy}, where the maximum improvement reaches 24.9\%. For \textit{Portability}, LGA based layer selection for editing Gemma3-12B exhibits consistently higher performance across all datasets. {On the other hand, LGA improves over SaLEM by $+17.53\%$ in \textit{Rewrite}, $+11.63\%$ in \textit{Rephrase}, $+24.36\%$ in \textit{Locality}, $+7.63\%$ in \textit{Portability}, and $+5.11\%$ in \textit{Fluency} averaged across all models and datasets.} Overall, these results indicate that LGA maintains improved performance across metrics, with observable improvements over CMA.\vspace{1.5mm}

\begin{figure}[t]
    \centering
    \vspace{4mm}
    \includegraphics[width=1\textwidth]{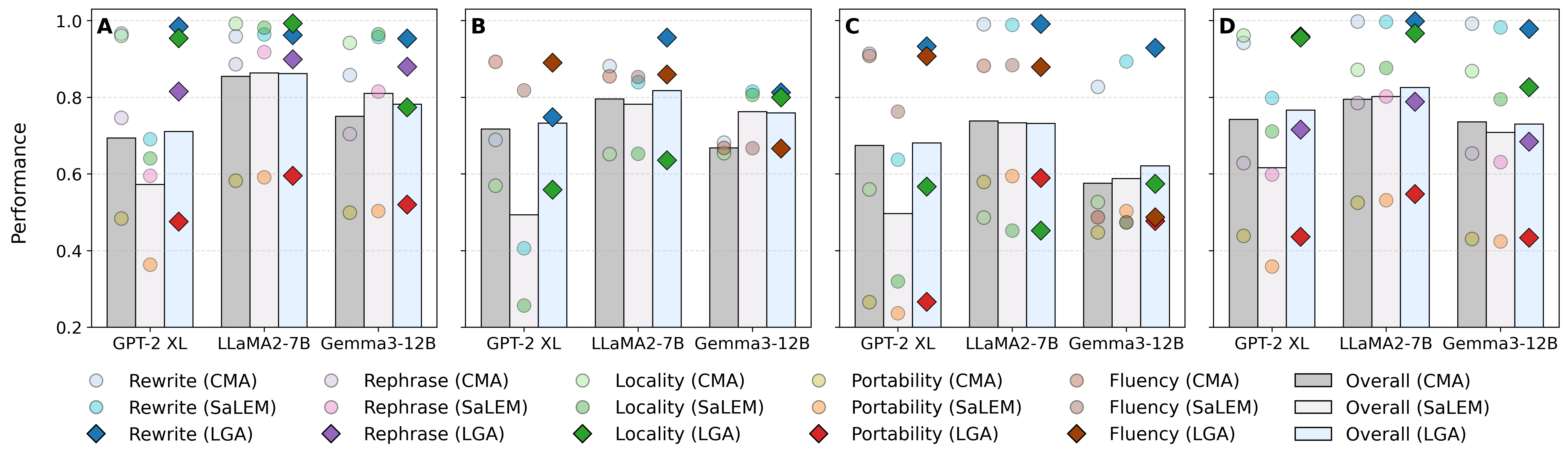}\vspace{-1mm}
    \caption{{Performance comparison between LGA, SaLEM, and CMA across different LLMs and the (A) \textit{ZSRE}, (B) \textit{WikiBio}, (C) \textit{WikiCounterfact}, and (D) \textit{Counterfact} datasets. Overall, LGA outperforms both CMA and SaLEM, attaining improved knowledge editing performance.}}\vspace{-4mm}
    \label{fig:lga-cma}\vspace{-0mm}
\end{figure}

\textbf{Comparison with Random Layer Selection.}
Next, to verify that LGA's layer selection criterion is informative rather than arbitrary, we compare LGA-selected layers with randomly selected layers across three models on the \textit{ZSRE} dataset as presented in Table~\ref{tab:random_selection_comparison_main}. The random baseline uniformly samples three layers, and we report its mean performance. LGA consistently outperforms random selection across all three models, achieving an average \textit{Overall} performance gain of 8.69\%.

\begin{wrapfigure}{r}{0.55\linewidth}
\vspace{-5.5mm}
\centering
\includegraphics[width=\linewidth]{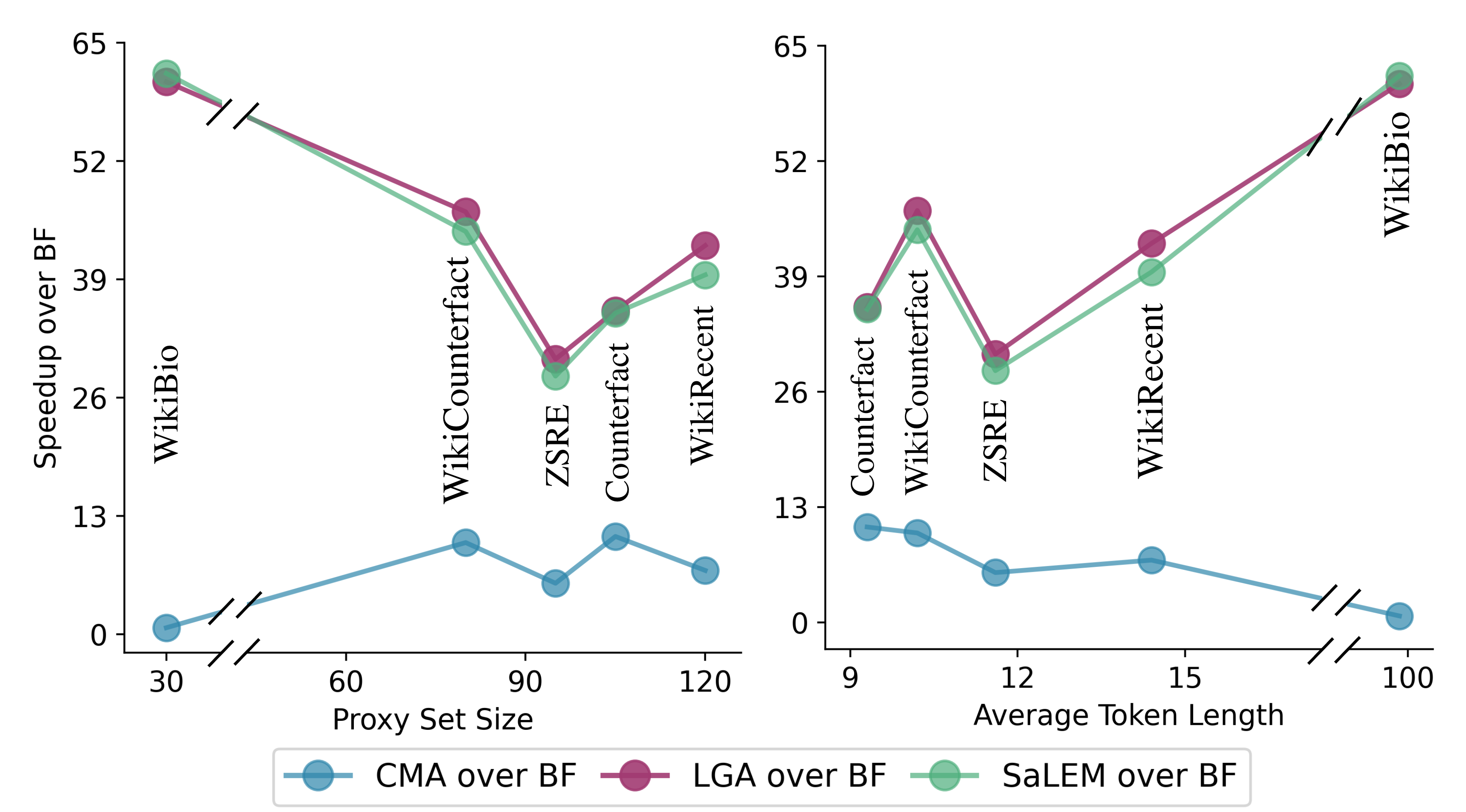}\vspace{-2mm}
\caption{{\looseness-1 Runtime analysis of LGA, CMA, and SaLEM over Brute-Force (BF) golden layer search for editing via R-ROME on GPT-2 XL. Each of the five datasets: \textit{ZSRE}, \textit{WikiBio}, \textit{WikiCounterfact}, \textit{WikiRecent}, and \textit{Counterfact}, are categorized in terms of the average query token length (\textit{left}) and the proxy size (\textit{right}). LGA is extremely computationally efficient and yet attains improved editing performance compared to baselines.}}
\label{fig:runtime_analysis}
\vspace{-12pt}
\end{wrapfigure}

\textbf{Computational Efficiency Analysis.}
\looseness-1We conduct experiments to assess the computational efficiency benefits of LGA. In Figure~\ref{fig:runtime_analysis}, we compare the runtime of LGA, CMA, and SaLEM over \textit{Brute-Force (BF)} golden layer search for the R-ROME knowledge editing method on GPT-2 XL. While the runtime of all methods generally increases with longer input sequences and larger proxy sizes, LGA and SaLEM achieve substantial speedups over both CMA and BF across all datasets, demonstrating exceptional computational efficiency. In contrast, CMA remains relatively inefficient and, in some cases, exhibits runtime comparable to BF, particularly on \textit{WikiBio}, which possesses high average token length. Quantitatively, LGA attains consistent speedups ranging from approximately $30\times$ to over $60\times$ relative to BF, while CMA has a lower bound of $1\times$. While LGA and SaLEM attain similar runtime performance, LGA leads to consistent gains in editing performance (as our results from previous sections show). These results indicate that LGA scales efficiently with dataset characteristics and proxy set size while attaining improved editing performance in comparison to CMA and LGA. \vspace{1mm}

\looseness-1 \textbf{Additional Editing Settings and Miscellaneous Results.}
We also conduct additional experiments with other knowledge editing settings and miscellaneous ablations to demonstrate the efficacy of LGA over baselines. First, in Appendix~\ref{app:sequential_editing_comparison}, we explore the \textit{sequential editing} task \cite{zhang2024comprehensive} and find the layer selected by LGA maintains stronger performance under multiple consecutive edits compared to CMA and SaLEM. Furthermore, in Appendix~\ref{app:composite_longform_unstructured} we study \textit{long-form} \cite{rosati-etal-2024-long}, \textit{compositional} \cite{ma2024neighboring}, and \textit{unstructured} \cite{deng2025everything} \textit{editing}, where LGA again achieves consistently improved performance.
We next provide ablation results in Appendix~\ref{app:random_comparison} demonstrating that CMA serves as a stronger baseline than random layer selection, and LGA outperforms both.
Finally, since SaLEM and LGA both utilize gradient information for layer selection, we also undertake deeper analysis in Appendix~\ref{app:l2norm_comparison} that shows how LGA utilizes cross-knowledge interactions for improved layer selection.

\vspace{-0.75mm}
\section{Conclusion} \vspace{-0.75mm}
In this paper, we studied the knowledge editing task in LLMs, which seeks to update the model's output for a given query to new target knowledge, without impacting the other desirable knowledge learned by the model during pre-training. Generally, knowledge editing methods first identify a specific layer to edit using standard approaches such as CMA and SaLEM, and then perform a minimal parameter update at that layer. Motivated by prior work demonstrating the inefficacy of CMA at selecting the top-performing editing layers, we analyzed editing performance across layers and found evidence for the existence of \textit{golden layers} that achieve on average, near-optimal or statistically similar editing performance compared to sample-wise optimal layers. We then proposed \textit{Layer Gradient Analysis (LGA)}, a novel and computationally efficient gradient-based strategy for robustly estimating these golden layers using a given proxy set of queries. Through several experiments across various benchmark datasets, LLMs, and editing methods, we demonstrated the significant gains achieved by LGA over current layer-selection method baselines, in terms of both editing performance and computational efficiency.

\section*{Acknowledgments}
We gratefully acknowledge the support of the Google TPU Builders Program, which provided support and access to computational resources, and Google Tunix library for post-training functionality and flexibility to enable this work.

\bibliographystyle{plainnat}
\bibliography{refs}

\clearpage

%%%%%%%%%%%%%%%%%%%%%%%%%%%%%%%%%%%%%%%%%%%%%%%%%%%%%%%%%%%%%%%%%%%%%%%%%%%%%%%
%%%%%%%%%%%%%%%%%%%%%%%%%%%%%%%%%%%%%%%%%%%%%%%%%%%%%%%%%%%%%%%%%%%%%%%%%%%%%%%
% APPENDIX
%%%%%%%%%%%%%%%%%%%%%%%%%%%%%%%%%%%%%%%%%%%%%%%%%%%%%%%%%%%%%%%%%%%%%%%%%%%%%%%
%%%%%%%%%%%%%%%%%%%%%%%%%%%%%%%%%%%%%%%%%%%%%%%%%%%%%%%%%%%%%%%%%%%%%%%%%%%%%%%

\appendix
\section*{Appendix}

\section{Intuition and Justification of the LGA Criterion}\label{app:lga_justification}

\subsection{Intuition of LGA Definition}\label{app:lga_intuition}
We put forward the intuition behind why maximizing the proposed impact of both new and old knowledge in LGA helps in identifying the golden layer:
\begin{itemize}[nosep, leftmargin=2em]
    \item A related study on memorization \cite{stoehr2024localizing} states that memorized examples, i.e., old knowledge, can be unlearned by fine-tuning only the high-gradient weights of these samples. Hence, LGA selects the layers most impacted by the old knowledge to achieve an unlearning effect on the old knowledge.
    \item Surgical Fine-Tuning \cite{lee2023surgical} selects layers by their target-task (analogous to new knowledge) gradient magnitude on the target task, on the hypothesis that ``layers with large gradient magnitudes may carry more information about the target task than others and can therefore be more useful''. Thus, LGA selects the layers most impacted by the new knowledge to maximize the learning of the new task.
    \item Finally, by definition, LGA aggregates the impact over the proxy samples (Section~\ref{sec:lga_approach}), and other studies \cite{meng2023memit, meng2022locatingRome} observe that strong causal mediation is limited, averaged across samples, to a fixed set of layers.
\end{itemize}

\subsection{Formal Justification of the LGA Criterion}\label{app:lga_derivation}
We provide a derivation showing that maximizing the LGA criterion minimizes a guaranteed upper bound on the required parameter-update magnitude. Essentially, we derive the minimum-norm weight update needed to achieve a post-edit first-order reduction in the new-knowledge loss. We then bound the new-knowledge gradient using the LGA criterion via the Cauchy-Schwarz inequality. Combining these results yields an upper bound on the required parameter update that is inversely proportional to the LGA score.

\textbf{\textit{Definitions:}} Let $\theta$ be the pre-edit parameters, $\theta_L$ the layer-$L$ parameters, and $\Delta\theta_L$ the edit update. For query $Q$, original knowledge $K$, and new target $K'$, we define:
\begin{equation*}
g_L = \nabla_{\theta_L}\ell(\theta; Q\cup K), \qquad g'_L = \nabla_{\theta_L}\ell(\theta; Q\cup K'), \qquad \phi_L = g_L^\top g'_L,
\end{equation*}
where both gradients are evaluated at $\theta$, and assume $g'_L \neq 0$.

\textbf{\textit{Edit constraint:}} Using the first-order Taylor approximation, we have:
\begin{equation*}
\ell(\theta + \Delta\theta_L;\,Q\cup K') \approx \ell(\theta;\,Q\cup K') + (g'_L)^\top\Delta\theta_L.
\end{equation*}
Requiring the new-knowledge loss to decrease by $\delta>0$ gives us:
\begin{equation}\label{eq:lga_C}
(g'_L)^\top\Delta\theta_L = -\delta. \tag{C}
\end{equation}

\textbf{\textit{Minimum-norm update:}} Also consider the optimization:
\begin{equation*}
\min_{\Delta\theta_L}\ \lVert\Delta\theta_L\rVert \quad \text{subject to} \quad (g'_L)^\top\Delta\theta_L = -\delta.
\end{equation*}
By Cauchy--Schwarz inequality, we have:
\begin{equation*}
\delta = |(g'_L)^\top\Delta\theta_L| \le \lVert g'_L\rVert\lVert\Delta\theta_L\rVert \quad\Longrightarrow\quad \lVert\Delta\theta_L\rVert \ge \frac{\delta}{\lVert g'_L\rVert}.
\end{equation*}
The equality holds only when the two are collinear. The smallest feasible update is therefore a scalar multiple of the new-knowledge gradient, $\Delta\theta_L = c\, g'_L$ for some scalar $c$. Substituting this into \eqref{eq:lga_C} determines $c$:
\begin{equation}\label{eq:lga_minnorm}
\Delta\theta_L = -\frac{\delta}{\lVert g'_L\rVert^2}g'_L, \qquad \lVert\Delta\theta_L\rVert = \frac{\delta}{\lVert g'_L\rVert}.
\end{equation}

\textbf{\textit{New-gradient norm bound:}} Applying Cauchy--Schwarz now to the LGA criterion, we have:
\begin{equation*}
\phi_L = g_L^\top g'_L \le \lVert g_L\rVert\lVert g'_L\rVert.
\end{equation*}
Thus, we can state:
\begin{equation}\label{eq:lga_newbound}
\lVert g'_L\rVert \ge \frac{\phi_L}{\lVert g_L\rVert}.
\end{equation}

\textbf{\textit{Uniform layer bound:}} Let $G = \max_L \lVert g_L\rVert$. Since $\lVert g_L\rVert \le G$:
\begin{equation}\label{eq:lga_uniform}
\lVert g'_L\rVert \ge \frac{\phi_L}{G}.
\end{equation}

\textbf{\textit{Target relation:}} Substituting \eqref{eq:lga_uniform} into \eqref{eq:lga_minnorm} gives us:
\begin{equation}\label{eq:lga_target}
\lVert\Delta\theta_L\rVert \le \frac{\delta G}{\phi_L}.
\end{equation}

\textbf{Maximizing the LGA Criterion Minimizes the Weight-Update Upper Bound.} LGA favors layers where the new knowledge can be incorporated with a smaller guaranteed weight update, thereby better preserving unrelated knowledge. From \eqref{eq:lga_target}, maximizing $\phi_L$ minimizes a guaranteed upper bound on the update required to install the new target.

\textbf{Relationship Between the LGA Criterion and the Knowledge Gradient Magnitude.} We can show that maximizing the LGA criterion requires a contribution from both gradients, rather than either one alone. From \eqref{eq:lga_newbound}, and symmetrically:
\begin{equation*}
\lVert g'_L\rVert \ge \frac{\phi_L}{\lVert g_L\rVert}, \qquad \lVert g_L\rVert \ge \frac{\phi_L}{\lVert g'_L\rVert}.
\end{equation*}
Maximizing $\phi_L$ therefore raises the guaranteed lower bound on each gradient magnitude relative to the other. That is, neither gradient can be small relative to its counterpart when the criterion is large. A positive $\phi_L$ additionally requires the two gradients to be aligned.

This joint dependence motivates using both gradients in LGA. This is also reflected empirically, where maximizing only one of the two gradients is not sufficient for effective editing. LGA($K'$), which uses only the new-knowledge gradient, comparatively underperforms against LGA (Table~\ref{tab:l2norm_comparison}).

\section{Implementation Details}\label{app:implementation}
Here we introduce the implementation details for knowledge editing in terms of datasets, models, editing methods, and editing performance metrics.
Note that similar to CMA and other prior work in knowledge editing, we focus only on the MLP layer modules \cite{meng2022locatingRome, geva2021transformer} for layer selection/editing for both LGA and CMA. Furthermore, for our LGA method, in accordance with recent work demonstrating that outlier gradients comprise instances that are detrimental to training loss \cite{chhabra2025oga, bejan2023make} we exclude layers with overall outlying gradient scores (measured using Tukey's fences and the interquartile range \cite{hoaglin1986performance}. Unless otherwise specified, we fix the Tukey constant to 1 across all evaluation tasks). Finally, we build LGA upon the EasyEdit knowledge editing framework \cite{wang2024easyediteasytouseknowledgeediting} to ensure consistency in performance and standardized evaluation across all LLMs, datasets, and editing methods.

\subsection{Datasets and LLMs}
We conduct experiments on five datasets that are popularly used in prior editing work: \textit{ZSRE} \cite{levy2017zero, wang2024easyediteasytouseknowledgeediting}, \textit{WikiBio} \cite{zhang2024comprehensive, hartvigsen2023aging}, \textit{WikiCounterfact} \cite{zhang2024comprehensive}, \textit{WikiRecent} \cite{zhang2024comprehensive, cohen2024evaluating}, and \textit{Counterfact} \cite{meng2022locatingRome}. \textit{ZSRE}, \textit{Counterfact}, and \textit{WikiCounterfact} are primarily used to test the introduction of unfactual knowledge, whereas \textit{WikiBio} and \textit{WikiCounterfact} focus on correcting existing knowledge. For each dataset, we use approximately 10\% of samples as the proxy set and the remainder 90\% as the test set; for example, resulting in approximately 100/1000, 30/270, 100/1000, and 120/1080 samples for the proxy/test sets of \textit{ZSRE}, \textit{WikiBio}, \textit{WikiCounterfact}, and \textit{WikiRecent}, respectively.
Moreover, we experiment with three very different LLMs in our paper: GPT-2 XL \cite{radford2019language}, LLaMA2-7B \cite{touvron2023llama}, and Gemma3-12B \cite{team2025gemma}.

\subsection{Editing Methods}
We primarily use the R-ROME \cite{gupta2024rebuilding} editing method in our experiments due to its superlative editing performance across metrics. This choice is further motivated by its design as an extension of ROME that more efficiently supports sequential edits. To assess the generality of our approach across different editing methods, we additionally evaluate on other editing methods, such as EMMET \cite{gupta2024emmet} and ROME \cite{meng2022locatingRome}.

% \subsection{Performance Metrics}
% We evaluate editing performance using five commonly used metrics: \textit{Rewrite Accuracy}, \textit{Rephrase Accuracy}, \textit{Locality}, \textit{Portability}, and \textit{Fluency}. \textit{Rewrite Accuracy} measures whether the edit succeeds for the original query associated with the edited knowledge. \textit{Rephrase Accuracy} evaluates the model’s ability to recall the edited knowledge under semantically equivalent, rephrased queries. \textit{Locality} assesses whether the edit unintentionally alters model outputs for unrelated queries \cite{meng2022locatingRome, wang2024easyediteasytouseknowledgeediting, yao2023editinglargelanguagemodels, zhang2024comprehensive}. \textit{Portability} quantifies the model’s ability to generalize the edited knowledge to downstream reasoning tasks \cite{yao2023editinglargelanguagemodels}. \textit{Fluency} evaluates the linguistic quality of the generated responses. Not all datasets support all metrics \cite{wang2024easyediteasytouseknowledgeediting, zhang2024comprehensive}. All metrics indicate better editing performance for higher values. We additionally report an \textit{Overall} score computed as the average of all normalized metrics.

\subsection{Performance Metrics}\label{app:metrics}
We evaluate editing performance using five commonly used metrics: \textit{Rewrite Accuracy}, \textit{Rephrase Accuracy}, \textit{Locality}, \textit{Portability}, and \textit{Fluency}, following the evaluation protocol of \citet{zhang2024comprehensive} as implemented in EasyEdit \cite{wang2024easyediteasytouseknowledgeediting}. Not all datasets support all metrics \cite{wang2024easyediteasytouseknowledgeediting, zhang2024comprehensive}, and all metrics indicate better editing performance for higher values. Each metric is computed per edit request and averaged over the test set.

\textbf{Rewrite Accuracy.} measures whether the post-edit model produces the new target knowledge when queried with the edit prompt \cite{meng2022locatingRome, zhang2024comprehensive}. The target is scored under teacher forcing: at each target token position, the model's greedy prediction is compared to the corresponding target token, and the score is the fraction of matching tokens.

\textbf{Rephrase Accuracy.} measures whether the edit generalizes to a semantically equivalent paraphrase of the edit prompt \cite{yao2023editinglargelanguagemodels, zhang2024comprehensive}. It is computed identically to \textit{Rewrite Accuracy}, but on the paraphrased prompt.

\textbf{Locality.} measures whether knowledge unrelated to the edit remains unaffected \cite{meng2022locatingRome, yao2023editinglargelanguagemodels, zhang2024comprehensive}. Following \citet{zhang2024comprehensive}, it is computed over out-of-scope prompts as the fraction of token positions at which the post-edit model's prediction matches the pre-edit model's prediction. These prompts are unrelated questions for \textit{ZSRE}, neighborhood prompts about other subjects for \textit{Counterfact}, relation-specificity prompts for \textit{WikiBio}, and relation-specificity and forgetfulness prompts for \textit{WikiCounterfact} and \textit{WikiRecent}.

\textbf{Portability.} measures whether the edit generalizes beyond the edit prompt to related queries \cite{yao2023editinglargelanguagemodels, zhang2024comprehensive}. It is computed identically to \textit{Rewrite Accuracy}, but on one-hop reasoning questions for \textit{ZSRE} and \textit{Counterfact}, and on subject-aliasing, reasoning, and logical-generalization questions for \textit{WikiCounterfact} and \textit{WikiRecent}, averaged across the available question types.

\textbf{Fluency.} evaluates the generative capacity of the post-edit model \cite{meng2022locatingRome, zhang2024comprehensive}. It is computed as the weighted average of the bi-gram and tri-gram entropies of the text generated from the edit prompt, where a lower value indicates more repetitive text. Since this entropy does not lie in $[0,1]$, we min-max normalize it over all edit requests and layers of a given model and dataset.

Finally, we report an \textbf{Overall} score as the average of all metrics supported by a dataset, each min-max normalized over all edit requests and layers of a given model and dataset.

\section{Results on Additional Datasets}\label{app:addn_results}
\subsection{Proxy-Selected versus Test-Selected Layer Editing Performances for \textit{WikiBio} and \textit{Counterfact}}\label{app:proxy_vs_test}

Figure~\ref{fig:proxy_vs_test-appendix} presents the knowledge editing performance on GPT-2 XL with R-ROME editing method when using golden layers identified on a proxy set versus a test set on \textit{WikiBio} and \textit{{WikiRecent}}. We observe that the editing performance of the golden layers identified from the proxy set and the test set remains very close, similar to trends observed in the main paper.

\begin{figure}[htbp]
    \centering
    \includegraphics[width=0.7\textwidth]{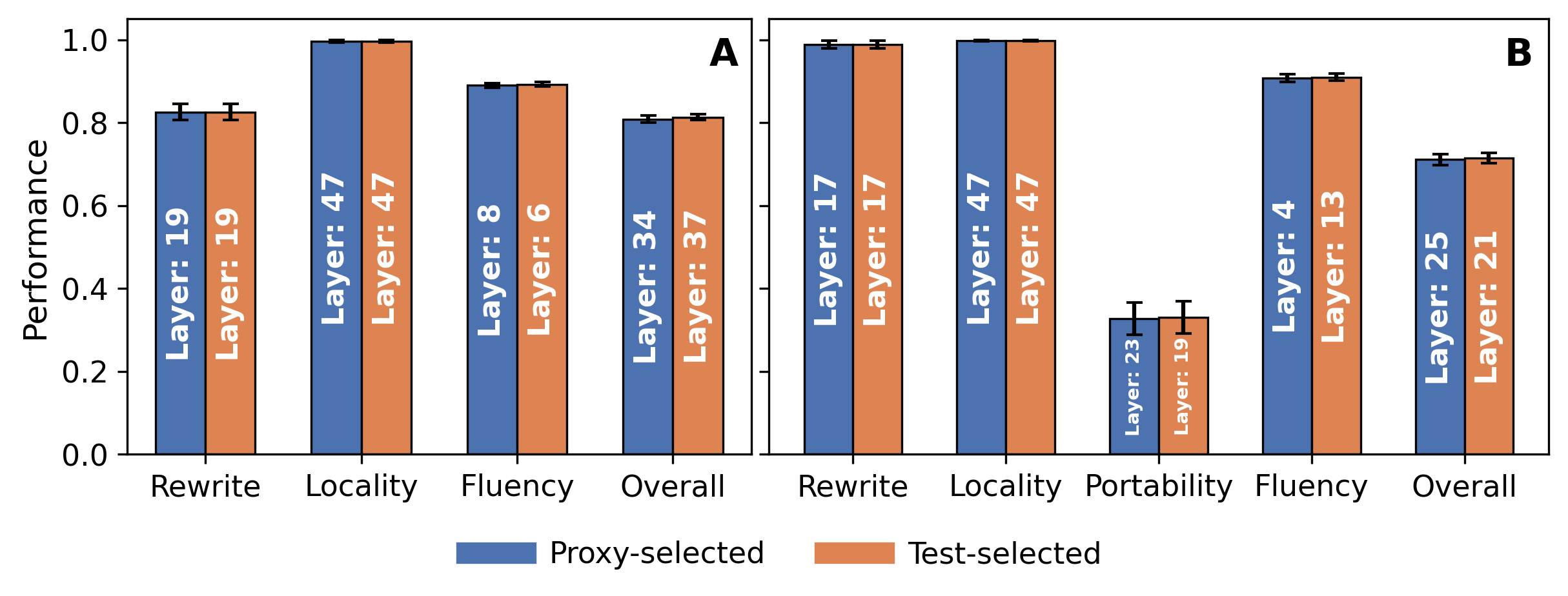}
    \caption{Performance of golden layers selected via the proxy and test sets with GPT-2 XL on (A) \textit{WikiBio} and (B) \textit{{WikiRecent}}. Editing performance evaluation is conducted on the test set queries.}
    \label{fig:proxy_vs_test-appendix}
\end{figure}

\subsection{Proxy Golden Layer versus Test Golden Layer on \textit{ZSRE}, \textit{WikiBio}, \textit{WikiCounterfact}, \textit{WikiRecent}, \textit{Counterfact}}\label{app:proxy-vs-test-others}

Tables~\ref{tab:proxy_vs_test_rwa}, \ref{tab:proxy_vs_test_rpa}, \ref{tab:proxy_vs_test_loc}, \ref{tab:proxy_vs_test_port}, \ref{tab:proxy_vs_test_flc}, and \ref{tab:proxy_vs_test_ov}, report \textit{Rewrite Accuracy}, \textit{Rephrase Accuracy}, \textit{Locality}, \textit{Portability}, \textit{Fluency}, and \textit{Overall} metrics evaluated on the test set of the Proxy Optimal Layer and the Golden Layer under R-ROME editing across \textit{ZSRE}, \textit{WikiBio}, \textit{WikiCounterfact}, \textit{WikiRecent}, and \textit{Counterfact} datasets and across three LLMs: GPT-2 XL, LLaMA2-7B, and Gemma3-12B. We observe that the golden layers identified from the proxy set and the test set often coincide. The resulting editing performance remains largely consistent, even when the selected golden layers differ.
% \sd{Describe results at a high-level}

\begin{table}[htbp]
\centering
\small
\setlength{\tabcolsep}{2pt}
\caption{The \textit{Rewrite Accuracy} of the Proxy Optimal Layer (top performing layer in the proxy set) and Golden Layer (top performing layer in the test set), for each of the editing performance metrics on the \textit{ZSRE}, \textit{WikiBio}, \textit{WikiCounterfact}, \textit{WikiRecent}, and \textit{Counterfact} datasets on three LLMs, GPT-2 XL, LLaMA2-7B, and Gemma3-12B editing with R-ROME. The optimal layer selection is based on \textit{Rewrite Accuracy} evaluated on the test set.}
\begin{tabular}{ c | c |c c | c c | c c | c c | c c}
\toprule
\multirow{2}{*}{Model} & \multirow{2}{*}{Set} 
& \multicolumn{2}{c|}{\textit{ZSRE}} 
& \multicolumn{2}{c|}{\textit{WikiBio}} 
& \multicolumn{2}{c|}{\textit{WikiCounterfact}} 
& \multicolumn{2}{c|}{\textit{WikiRecent}} 
& \multicolumn{2}{c}{\textit{Counterfact}} \\
\cmidrule{3-12}
& 
& Layer & Rewrite
& Layer & Rewrite
& Layer & Rewrite
& Layer & Rewrite
& Layer & Rewrite \\
\midrule
\multirow{2}{*}{\rotatebox{0}{GPT-2 XL}}&Proxy & 15 & 0.9962 & 19 & 0.8256 & 12 & 0.9380 & 17 & 0.9887 & 16 & 0.9796\\
        &Test & 19 & 0.9993 & 19 & 0.8256 & 15 & 0.9537 & 17 & 0.9887 & 18 & 0.9871\\
\midrule
\multirow{2}{*}{\rotatebox{0}{LLaMA2-7B}}&Proxy & 12 & 0.9650 & 18 & 0.9835 & 3 & 0.9888 & 12 & 0.9783 & 2 & 0.9969\\
        &Test & 10 & 0.9667 & 13 & 0.9912 & 12 & 0.9922 & 19 & 0.9792 & 4 & 0.9995\\
\midrule
\multirow{2}{*}{\rotatebox{0}{Gemma3-12B}}&Proxy & 27 & 0.9328 & 13 & 0.8538 & 19 & 0.9340 & 11 & 0.9784 & 8 & 0.9941\\
        &Test & 18 & 0.9763 & 13 & 0.8538 & 13 & 0.9415 & 19 & 0.9789 & 35 & 1.0000\\
        \bottomrule
\end{tabular}
\label{tab:proxy_vs_test_rwa}
\end{table}

\begin{table}[htbp]
\centering
\small
\caption{\looseness=-1The \textit{Rephrase Accuracy} of the Proxy Optimal Layer (top performing layer in the proxy set) and Golden Layer (top performing layer in the test set), for each of the editing performance metrics on the \textit{ZSRE} and \textit{Counterfact} datasets on three LLMs, GPT-2 XL, LLaMA2-7B, and Gemma3-12B editing with R-ROME. The optimal layer selection is based on \textit{Rephrase Accuracy} evaluated on the test set.}
\setlength{\tabcolsep}{6pt}
\begin{tabular}{ c | c | c c | c c}
\toprule
\multirow{2}{*}{Model} & \multirow{2}{*}{Set} 
& \multicolumn{2}{c|}{\textit{ZSRE}} 
& \multicolumn{2}{c}{\textit{Counterfact}} \\
\cmidrule{3-6}
& 
& Layer & Rephrase
& Layer & Rephrase \\
\midrule
\multirow{2}{*}{\rotatebox{0}{GPT-2 XL}}&Proxy & 16 & 0.8823 & 18 & 0.8034\\
        &Test & 17 & 0.8838 & 18 & 0.8034\\
\midrule
\multirow{2}{*}{\rotatebox{0}{LLaMA2-7B}}&Proxy & 7 & 0.9136 & 6 & 0.8023\\
        &Test & 6 & 0.9176 & 15 & 0.8196\\
\midrule
\multirow{2}{*}{\rotatebox{0}{Gemma3-12B}}&Proxy & 39 & 0.9108 & 38 & 0.9221\\
        &Test & 40 & 0.9294 & 40 & 0.9353\\
        \bottomrule
\end{tabular}
\label{tab:proxy_vs_test_rpa}
\end{table}

\begin{table}[htbp]
\centering
\small
\caption{\looseness=-1The \textit{Locality} of the Proxy Optimal Layer (top performing layer in the proxy set) and Golden Layer (top performing layer in the test set), for each of the editing performance metrics on the \textit{ZSRE}, \textit{WikiBio}, \textit{WikiCounterfact}, \textit{WikiRecent}, and \textit{Counterfact} datasets on three LLMs, GPT-2 XL, LLaMA2-7B, and Gemma3-12B editing with R-ROME. The optimal layer selection is based on \textit{Locality} evaluated on the test set.}
\setlength{\tabcolsep}{2pt}
\begin{tabular}{ c | c |c c | c c | c c | c c | c c}
\toprule
\multirow{2}{*}{Model} & \multirow{2}{*}{Set} 
& \multicolumn{2}{c|}{\textit{ZSRE}} 
& \multicolumn{2}{c|}{\textit{WikiBio}} 
& \multicolumn{2}{c|}{\textit{WikiCounterfact}} 
& \multicolumn{2}{c|}{\textit{WikiRecent}} 
& \multicolumn{2}{c}{\textit{Counterfact}} \\
\cmidrule{3-12}
& 
& Layer & Locality
& Layer & Locality
& Layer & Locality
& Layer & Locality
& Layer & Locality \\
\midrule
\multirow{2}{*}{\rotatebox{0}{GPT-2 XL}}&Proxy & 47 & 0.9959 & 46 & 0.9882 & 47 & 0.9965 & 47 & 0.9977 & 47 & 0.9968\\
        &Test & 47 & 0.9959 & 47 & 0.9964 & 47 & 0.9965 & 47 & 0.9977 & 47 & 0.9968\\
\midrule
\multirow{2}{*}{\rotatebox{0}{LLaMA2-7B}}&Proxy & 23 & 0.9895 & 23 & 0.8436 & 30 & 0.7402 & 25 & 0.6865 & 2 & 0.9747\\
        &Test & 3 & 0.9927 & 23 & 0.8436 & 30 & 0.7402 & 24 & 0.7016 & 2 & 0.9747\\
\midrule
\multirow{2}{*}{\rotatebox{0}{Gemma3-12B}}&Proxy & 11 & 0.9795 & 46 & 0.7870 & 46 & 0.7411 & 0 & 0.6523 & 0 & 0.9305\\
        &Test & 11 & 0.9795 & 15 & 0.8476 & 46 & 0.7411 & 46 & 0.7005 & 0 & 0.9305\\
\bottomrule
\end{tabular}
\label{tab:proxy_vs_test_loc}
\end{table}

\begin{table}[htbp]
\centering
\small
\caption{\looseness-1The \textit{Portability} of the Proxy Optimal Layer (top performing layer in the proxy set) and Golden Layer (top performing layer in the test set), for each of the editing performance metrics on the \textit{ZSRE}, \textit{WikiCounterfact}, \textit{WikiRecent}, and \textit{Counterfact} datasets on three LLMs, GPT-2 XL, LLaMA2-7B, and Gemma3-12B editing with R-ROME. The optimal layer selection is based on \textit{Portability} evaluated on the test set.}
\setlength{\tabcolsep}{4pt}
\begin{tabular}{ c | c |c c | c c | c c | c c}
\toprule
\multirow{2}{*}{Model} & \multirow{2}{*}{Set} 
& \multicolumn{2}{c|}{\textit{ZSRE}} 
& \multicolumn{2}{c|}{\textit{WikiCounterfact}} 
& \multicolumn{2}{c|}{\textit{WikiRecent}} 
& \multicolumn{2}{c}{\textit{Counterfact}} \\
\cmidrule{3-10}
& 
& Layer & Portability
& Layer & Portability
& Layer & Portability
& Layer & Portability \\
\midrule
\multirow{2}{*}{\rotatebox{0}{GPT-2 XL}}&Proxy & 23 & 0.4740 & 1 & 0.3032 & 23 & 0.3271 & 25 & 0.4392\\
        &Test & 29 & 0.4844 & 1 & 0.3032 & 19 & 0.3302 & 15 & 0.4450\\
\midrule
\multirow{2}{*}{\rotatebox{0}{LLaMA2-7B}}&Proxy & 11 & 0.5979 & 17 & 0.6154 & 8 & 0.5719 & 15 & 0.5336\\
        &Test & 12 & 0.5993 & 16 & 0.6199 & 16 & 0.5839 & 3 & 0.5476\\
\midrule
\multirow{2}{*}{\rotatebox{0}{Gemma3-12B}}&Proxy & 23 & 0.5442 & 43 & 0.6594 & 26 & 0.5806 & 1 & 0.4291\\
        &Test & 22 & 0.5480 & 42 & 0.6698 & 23 & 0.6103 & 0 & 0.4370\\
        \bottomrule
\end{tabular}
\label{tab:proxy_vs_test_port}
\end{table}

\begin{table}[htbp]
\centering
\small
\caption{The \textit{Fluency} of the Proxy Optimal Layer (top performing layer in the proxy set) and Golden Layer (top performing layer in the test set), for each of the editing performance metrics on the \textit{WikiBio}, \textit{WikiCounterfact}, and \textit{WikiRecent} datasets on three LLMs, GPT-2 XL, LLaMA2-7B, and Gemma3-12B editing with R-ROME. The optimal layer selection is based on \textit{Fluency} evaluated on the test set.}
\setlength{\tabcolsep}{5pt}
\begin{tabular}{ c | c |c c | c c | c c}
\toprule
\multirow{2}{*}{Model} & \multirow{2}{*}{Set} 
& \multicolumn{2}{c|}{\textit{WikiBio}} 
& \multicolumn{2}{c|}{\textit{WikiCounterfact}} 
& \multicolumn{2}{c}{\textit{WikiRecent}} \\
\cmidrule{3-8}
& 
& Layer & Fluency
& Layer & Fluency
& Layer & Fluency \\
\midrule
\multirow{2}{*}{\rotatebox{0}{GPT-2 XL}}&Proxy & 8 & 0.8897 & 45 & 0.9058 & 4 & 0.9081\\
        &Test & 6 & 0.8927 & 46 & 0.9157 & 13 & 0.9095\\
\midrule
\multirow{2}{*}{\rotatebox{0}{LLaMA2-7B}}&Proxy & 5 & 0.8595 & 6 & 0.8824 & 4 & 0.8800\\
        &Test & 6 & 0.8597 & 3 & 0.8837 & 4 & 0.8800\\
\midrule
\multirow{2}{*}{\rotatebox{0}{Gemma3-12B}}&Proxy & 32 & 0.7071 & 35 & 0.5824 & 32 & 0.6031\\
        &Test & 35 & 0.7180 & 32 & 0.5973 & 32 & 0.6031\\
\bottomrule
\end{tabular}
\label{tab:proxy_vs_test_flc}
\end{table}

\begin{table}[htbp]
\centering
\small
\caption{\looseness-1The \textit{Overall} of the Proxy Optimal Layer (top performing layer in the proxy set) and Golden Layer (top performing layer in the test set), for each of the editing performance metrics on the \textit{ZSRE}, \textit{WikiBio}, \textit{WikiCounterfact}, \textit{WikiRecent}, and \textit{Counterfact} datasets on three LLMs, GPT-2 XL, LLaMA2-7B, and Gemma3-12B editing with R-ROME. The optimal layer selection is based on \textit{Overall} evaluated on the test set.}
\setlength{\tabcolsep}{2pt}
\begin{tabular}{ c | c |c c | c c | c c | c c | c c}
\toprule
\multirow{2}{*}{Model} & \multirow{2}{*}{Set} 
& \multicolumn{2}{c|}{\textit{ZSRE}} 
& \multicolumn{2}{c|}{\textit{WikiBio}} 
& \multicolumn{2}{c|}{\textit{WikiCounterfact}} 
& \multicolumn{2}{c|}{\textit{WikiRecent}} 
& \multicolumn{2}{c}{\textit{Counterfact}} \\
\cmidrule{3-12}
& 
& Layer & Overall
& Layer & Overall
& Layer & Overall
& Layer & Overall
& Layer & Overall \\
\midrule
\multirow{2}{*}{\rotatebox{0}{GPT-2 XL}}&Proxy & 16 & 0.8286 & 34 & 0.8083 & 15 & 0.6916 & 25 & 0.7108 & 9 & 0.7726\\
        &Test & 16 & 0.8286 & 37 & 0.8133 & 17 & 0.6961 & 21 & 0.7139 & 16 & 0.7789\\
\midrule
\multirow{2}{*}{\rotatebox{0}{LLaMA2-7B}}&Proxy & 6 & 0.8638 & 16 & 0.8771 & 14 & 0.7656 & 13 & 0.7625 & 3 & 0.8255\\
        &Test & 7 & 0.8645 & 16 & 0.8771 & 14 & 0.7656 & 12 & 0.7640 & 3 & 0.8255\\
\midrule
\multirow{2}{*}{\rotatebox{0}{Gemma3-12B}}&Proxy & 19 & 0.8261 & 13 & 0.7780 & 41 & 0.6411 & 15 & 0.6474 & 9 & 0.7361\\
        &Test & 19 & 0.8261 & 15 & 0.7793 & 41 & 0.6411 & 21 & 0.6529 & 11 & 0.7386\\
\bottomrule
\end{tabular}
\label{tab:proxy_vs_test_ov}
\end{table}

\subsection{\textit{WikiRecent} Results Across Different Editing Methods and LLMs}
\label{app:counterfact_methods_llms}

\looseness-1 Table~\ref{tab:combined_tables_editing_methods_cf} presents the editing performance of three editing methods: R-ROME, EMMET, and ROME on GPT-2 XL for the \textit{WikiRecent} dataset. Notably, multiple layers often achieved identical performance scores; in such cases, we report results for the first tied layer. Table ~\ref{tab:combined_tables_editing_methods_llms} presents the editing performance of R-ROME editing method on three models GPT-2 XL, LLaMA2-7B, and Gemma3-12B for the \textit{WikiRecent} dataset. The table shows trends consistent with those reported in the main paper, with LGA-selected layers achieving significantly higher Overall editing performance than CMA across all datasets. LGA also consistently outperforms CMA on \textit{Rewrite Accuracy} and \textit{Rephrase Accuracy}, demonstrating robust improvements of our proposed approach in the case of \textit{WikiRecent} dataset, similar to the datasets presented in the main paper: \textit{ZSRE}, \textit{WikiBio}, \textit{WikiCounterfact}, and \textit{Counterfact}.
% \sd{You should describe both the table results at a high level and mention how the trends are similar as for the other datasets in the main paper and LGA achieves significant improvements over CMA}

\begin{table}[htbp]  
\centering
\caption{{Editing performance evaluation of LGA, SaLEM, and CMA across different knowledge editing methods and \textit{WikiRecent} dataset on GPT-2 XL. Different performance metrics are reported, where \textit{Rewrite Accuracy} $\rightarrow$ RwA,  \textit{Locality} $\rightarrow$ LOC, \textit{Portability} $\rightarrow$ PRT, \textit{Fluency} $\rightarrow$ FLC, and \textit{Overall} $\rightarrow$ OV. Layers are identified using the proxy set for both methods and evaluation is undertaken on an unseen test set.}}
\label{tab:combined_tables_editing_methods_cf}
\setlength{\tabcolsep}{4pt}
 \resizebox{0.65\textwidth}{!}{%
\begin{tabular}{c | c | c c c c c}
\toprule
Edit & Selection & RwA$_{(\uparrow)}$ & LOC$_{(\uparrow)}$ & PRT$_{(\uparrow)}$ & FLC$_{(\uparrow)}$ & OV$_{(\uparrow)}$ \\
\midrule

\multirow{3}{*}{R-ROME}
& CMA & {0.9622} & {0.6124} & {0.3038} & 0.9084 & 0.6967 \\
& SaLEM & {0.9828} & 0.6170 & {0.3192} & 0.9090 & {0.7070} \\
& LGA (Ours) & {0.9735} & 0.6058 & {0.3161} & {0.9048} & {0.7000} \\
\midrule
\multirow{3}{*}{EMMET}
& CMA & 0.9837  & 0.5029 & 0.2979 & 0.8568 & 0.6603 \\
& SaLEM & 0.9922  & 0.5485 & 0.3097 & 0.8475 & {0.6745} \\
& LGA (Ours) & 0.9892  & 0.5199 & 0.3044 & 0.8491 & 0.6656 \\
\midrule
\multirow{3}{*}{ROME}
& CMA & 0.9674  & 0.6105 & 0.3033 & 0.9106 & 0.6980 \\
& SaLEM & 0.9874  & 0.6136 & 0.3208 & 0.9089 & {0.7077} \\
& LGA (Ours) & 0.9773  & 0.6027 & 0.3149 & 0.9077 & 0.7007 \\

\bottomrule
\end{tabular}
 }
\end{table}

\begin{table}[htbp]  
\centering
\caption{{Editing performance evaluation of LGA, CMA, and SaLEM for \textit{WikiRecent} dataset and across three models: GPT-2 XL, LLaMA2-7B and Gemma3-12B with R-ROME. Different performance metrics are reported, where \textit{Rewrite Accuracy} $\rightarrow$ RwA,  \textit{Locality} $\rightarrow$ LOC, \textit{Portability} $\rightarrow$ PRT, \textit{Fluency} $\rightarrow$ FLC, and \textit{Overall} $\rightarrow$ OV. Layers are identified using the proxy set for both methods, and the evaluation is undertaken on an unseen test set.}}
\label{tab:combined_tables_editing_methods_llms}
\setlength{\tabcolsep}{4pt}
 \resizebox{0.65\textwidth}{!}{%
\begin{tabular}{c | c | c c c c c}
\toprule
Model & Selection & RwA & LOC & PRT & FLC & OV \\
\midrule

\multirow{3}{*}{GPT-2 XL}
& CMA & 0.9622 & 0.6124 & 0.3038 & 0.9084 & 0.6967 \\
& SaLEM & {0.9828} & 0.6170 & {0.3192} & 0.9090 & {0.7070} \\
& LGA (Ours) & 0.9735 & 0.6058 & 0.3161 & 0.9048 & {0.7000} \\
\midrule
\multirow{3}{*}{LLaMA2-7B}
& CMA & 0.9748 & 0.5629 & 0.5731 & 0.8718 & {0.7456} \\
& SaLEM & 0.9768 &	0.5478&	0.5769&	0.8708&	0.7431\\
& LGA (Ours) & 0.9750 & 0.5560 & 0.5699 & 0.8771 & 0.7445 \\
\midrule
\multirow{3}{*}{Gemma3-12B}
& CMA & 0.9710 & 0.5796 & 0.5216 & 0.5020 & 0.6436 \\
& SaLEM&0.9604 &	0.5166&	0.5558&	0.4884&	0.6303\\
& LGA (Ours) & 0.9775 & 0.5779 & 0.5327 & 0.5013 & {0.6474} \\

\bottomrule
\end{tabular}
}
\end{table}

\section{Sequential Editing}\label{app:sequential_editing_comparison}

For sequential editing experiments, we apply 10 edits sequentially, where the same randomly selected samples are used across datasets. This choice follows prior work showing that performance degradation becomes noticeable at around 10 sequential edits \cite{meng2023memit, zhang2024comprehensive}. Additionally, prior studies have also evaluated smaller sequential edit settings to analyze sequential editing behavior \cite{zhang2024comprehensive}, further motivating this setting. We present the performance of sequential edits in Table \ref{tab:sequential_editing_comparison}, where it is evident that LGA selected layers perform better compared to CMA and SaLEM selected layers.

\begin{table}[htbp]  
\centering
\caption{Sequential editing performance with 10 sequential edits for LGA, SaLEM, and CMA across three models: {GPT-2 XL}, {LLaMA2-7B}, and {Gemma3-12B} on five datasets: \textit{ZSRE}, \textit{WikiBio}, \textit{WikiCounterFact}, \textit{CounterFact}, and \textit{WikiRecent}. The same randomly selected edit samples are applied sequentially for each dataset. Different datasets support different performance metrics, where \textit{Rewrite Accuracy} $\rightarrow$ RwA, \textit{Rephrase Accuracy} $\rightarrow$ RpA, \textit{Locality} $\rightarrow$ LOC, \textit{Portability} $\rightarrow$ PRT, \textit{Fluency} $\rightarrow$ FLC, and \textit{Overall} $\rightarrow$ OV. Layers are identified using the proxy set for each method, and evaluation is conducted on unseen test data. Results demonstrate performance differences under sequential editing, with LGA-selected layers achieving a mean overall performance gain of 7.9\% and 511\% over CMA and SaLEM, respectively.}
\label{tab:sequential_editing_comparison}
\footnotesize
\begin{minipage}{0.5\textwidth}
\centering
\textit{{ZSRE}}
\medskip
{
\setlength{\tabcolsep}{1.04pt}
\resizebox{\linewidth}{!}{
\begin{tabular}{c | c | c c c c c}
\toprule
Edit & Selection & RwA & RpA & LOC & PRT & OV \\
\midrule

\multirow{3}{*}{\shortstack[c]{GPT-2 XL}}
& CMA & 0.5083 & 0.5083 & 0.6917 & 0.4583 &  0.5417 \\
& SaLEM & 0.1750 & 0.2083 & 0.0250 & 0.0833 &  0.1229 \\
& LGA & 0.8267 & 0.7817 & 0.7545 & 0.6117 &  {0.7436} \\
\midrule
\multirow{3}{*}{LLAMA2-7B}
& CMA & 0.9689 & 0.8822 & 1.0000 & 0.6862 &  0.8843 \\
& SaLEM & 0.9489 & 0.9489 & 0.8908 & 0.6198 &  0.8521 \\
& LGA & 0.9689 & 0.9689 & 0.9933 & 0.7364 &  {0.9169} \\
\midrule
\multirow{3}{*}{GEMMA3-12B}
& CMA & 0.6500 & 0.5917 & 0.6622 & 0.5833 &  {0.6218} \\
& SaLEM & 0.6106 & 0.5856 & 0.8847 & 0.6483 &  {0.6823} \\
& LGA & 0.4894 & 0.4117 & 0.6601 & 0.5167 &  0.5195 \\

\bottomrule
\end{tabular}
}
}
\end{minipage}%
\hfill
\begin{minipage}{0.5\textwidth}  % right column
\centering
{\textit{WikiBio}}
\medskip
{
\setlength{\tabcolsep}{3.3pt}
\resizebox{\linewidth}{!}{
\begin{tabular}{c | c | c c c c}
\toprule
Edit & Selection & RwA & LOC  & FLC  & OV  \\
\midrule

\multirow{3}{*}{GPT-2 XL}
& CMA & 0.5120 &  0.4042 &  0.8770 & {0.4581} \\
& SaLEM & 0.0865 &  0.0912 &  0.8368 & 0.0889 \\
& LGA & 0.5566 &  0.3542 &  0.9083 & 0.4554 \\
\midrule
\multirow{3}{*}{LLaMA2-7B}
& CMA & 0.8767 &  0.5217 &  0.6215 & 0.6992 \\
& SaLEM & 0.8882 & 0.6279 & 0.7436 & {0.7580} \\
& LGA & 0.9367 &  0.5158 &  0.6439 & {0.7263} \\
\midrule
\multirow{3}{*}{GEMMA3-12B}
& CMA & 0.7119 &  0.7336 &  0.6771 & 0.7075 \\
& SaLEM & 0.6574 &  0.7783 &  0.6157 & {0.7179} \\
& LGA & 0.8124 &  0.8001 &  0.6666 & {0.7597} \\

\bottomrule
\end{tabular}
}
}
\end{minipage}

\vspace{-1mm}  % small vertical space between rows

\begin{minipage}{0.50\textwidth}  % bottom-left
\centering
{\textit{WikiCounterFact}}  % width=0, centered
\medskip
{
\setlength{\tabcolsep}{1.04pt}
\resizebox{\linewidth}{!}{
\begin{tabular}{c | c | c c c c c}
\toprule
Edit & Selection & RwA  & LOC  & PRT & FLC  & OV  \\
\midrule

\multirow{3}{*}{GPT-2 XL}
& CMA & 0.7444 &  0.3983 & 0.3449 & 0.8768 & 0.4959 \\
& SaLEM & 0.0333 &  0.0014 & 0.0000 & 0.7531 & 0.0116 \\
& LGA & 0.9800 &  0.4877 & 0.4826 & 0.8769 & {0.6501} \\
\midrule
\multirow{3}{*}{LLaMA27B}
& CMA & 0.9857 &  0.4137 & 0.6217 & 0.8671 & 0.6737 \\
& SaLEM & 0.9857 &  0.5243 & 0.6571 & 0.9125 & {0.7224} \\
& LGA & 1.0000 &  0.4126 & 0.6907 & 0.9146 & {0.7011} \\
\midrule
\multirow{3}{*}{GEMMA3-12B}
& CMA & 0.5278 &  0.5011 & 0.3534 & 0.4155 & 0.4608 \\
& SaLEM & 0.3556 &  0.4712 & 0.2789 & 0.4243 & 0.3685 \\
& LGA & 0.4806 &  0.6750 & 0.4196 & 0.4143 & {0.5250} \\

\bottomrule
\end{tabular}
}
}
\end{minipage}%
\hfill
\begin{minipage}{0.5\textwidth}  % bottom-right
\centering
{\textit{Counterfact}}  % width=0, centered
\medskip
{
\setlength{\tabcolsep}{1.04pt}
\resizebox{\linewidth}{!}{
\begin{tabular}{c | c | c c c c c}
\toprule
Edit & Selection & RwA & RpA  & LOC  & PRT & OV \\
\midrule
\multirow{3}{*}{GPT-2 XL}
& CMA & 0.5000 & 0.4000 & 0.3000 & 0.2350 &  {0.3588} \\
& SaLEM & 0.0000 & 0.1000 & 0.0000 & 0.0500 &  0.0375 \\
& LGA & 1.0000 & 0.4000 & 0.3000 & 0.2267 &  {0.4817} \\
\midrule
\multirow{3}{*}{LLaMA2-7B}
& CMA & 1.0000 & 0.9000 & 0.4500 & 0.2250 &  0.6438 \\
& SaLEM & 1.0000 & 0.9000 & 0.3500 & 0.2583 &  0.6271 \\
& LGA & 1.0000 & 0.8000 & 0.9000 & 0.3083 &  {0.7521} \\
\midrule
\multirow{3}{*}{GEMMA3-12B}
& CMA & 1.0000 & 0.6000 & 0.9000 & 0.3000 &  {0.7000} \\
& SaLEM & 0.9000 & 0.5000 & 0.7000 & 0.3000 &  0.6000 \\
& LGA & 0.9000 & 0.5000 & 0.8000 & 0.3000 &  0.6250 \\

\bottomrule
\end{tabular}
}
}
\end{minipage}
\hfill
\begin{minipage}{0.5\textwidth}  % bottom-right
\centering
{\textit{WikiRecent}}  % width=0, centered
\medskip
{
\setlength{\tabcolsep}{1.04pt}
\resizebox{\linewidth}{!}{
\begin{tabular}{c | c | c c c c c}
\toprule
Edit & Selection & RwA & RpA & LOC & PRT & OV \\
\midrule
\multirow{3}{*}{GPT-2 XL}
& CMA & 0.8130 &  0.6465 & 0.3522 & 0.8885 & 0.6039 \\
& SaLEM & 0.8702 &  0.5677 & 0.3460 & 0.9298 & 0.5946 \\
& LGA & 0.8759 &  0.6559 & 0.3431 & 0.8492 & {0.6250} \\
\midrule
\multirow{3}{*}{LLaMA2-7B}
& CMA & 0.9500 &  0.6109 & 0.6145 & 0.8115 & {0.7251} \\
& SaLEM & 0.8708 &  0.6491 & 0.4976 & 0.9009 & 0.6725 \\
& LGA & 0.9333 &  0.6209 & 0.5769 & 0.7983 & 0.7104 \\
\midrule
\multirow{3}{*}{GEMMA3-12B}
& CMA & 0.9333 &  0.6894 & 0.5760 & 0.5673 & {0.7329} \\
& SaLEM & 0.7238 &  0.7073 & 0.4219 & 0.3157 & 0.6177 \\
& LGA & 0.9071 &  0.6142 & 0.5030 & 0.4689 & 0.6748 \\

\bottomrule
\end{tabular}
}
}
\end{minipage}
\vspace{-4mm}
\end{table}

\section{Composite, Longform, and Unstructured Editing}\label{app:composite_longform_unstructured}

\subsection{Composite Editing}
We evaluate compositional model editing on the \textit{PeakCF} dataset, following the recent protocols while reporting SRS (compositional success rate), SR-1 (single-edit success rate), GSR (generalization success rate), and LSR (Rouge-L-based generation score) as evaluation metrics \cite{piao2026ae, ma2024neighboring}. Table~\ref{tab:composite_editing} presents the performance of layers selected by LGA, SaLEM, and CMA. LGA achieves improvements of 2.38\% and 49.30\% over CMA and SaLEM, respectively, in terms of average performance. Layer selection is performed using a proxy set of size 30, and evaluation is conducted on a test set of size 270 randomly selected from the \textit{PeakCF} dataset. For the task of composite editing, Tukey constant is fixed to 0.5.

\begin{table}[htbp]
\centering
\small
\setlength{\tabcolsep}{2.5pt}
\caption{Compositional editing performance of LGA-selected layers, SaLEM, and CMA on the \textit{PeakCF} dataset across three models: {GPT-2 XL}, {LLaMA2-7B}, and {Gemma3-12B}. We use R-ROME for editing and evaluate using SRS, SR-1, GSR, and LSR metrics. Layer selection uses a proxy set of size 30, and evaluation is performed on 270 randomly sampled test examples from the PeakT dataset. Layers selected by LGA achieve average overall performance improvements of 2.38\% and 49.30\% over CMA and SaLEM, respectively.}
\resizebox{0.6\textwidth}{!}{
\begin{tabular}{ c | c | c c c c c}
\toprule
Model & Method & SRS $\uparrow$ & SR-1 $\uparrow$ & GSR $\uparrow$ & LSR $\uparrow$ & Overall $\uparrow$ \\
% \midrule

\midrule
\multirow{3}{*}{\shortstack[c]{GPT-2 XL}}
& CMA & 0.4562 & 0.1630 & 0.4023 & 0.0105 & {0.3462} \\
& SaLEM & 0.2663 &  0.1556 &  0.2073 &  0.0154 & 0.1611 \\
& LGA & 0.5513 & 0.4278 & 0.4357 & 0.0197 & \textbf{0.3586} \\
\midrule
\multirow{3}{*}{LLAMA2-7B}
& CMA & 0.6417 &  0.4944 &   0.5297  & 0.0282 & {0.4235} \\
& SaLEM & 0.6417 &  0.4944 &   0.5297  & 0.0282 & {0.4235} \\
& LGA & 0.6513 & 0.4926 & 0.5398 & 0.0290 & \textbf{0.4282} \\
\midrule
\multirow{3}{*}{GEMMA3-12B}
& CMA & 0.3986 & 0.2593 &  0.3150  & 0.0214 & {0.2485} \\
& SaLEM & 0.3417 &   0.1741  &  0.2845 &  0.0200 & {0.2050} \\
& LGA  & 0.3939 & 0.2889 & 0.3103 & 0.0252 & \textbf{0.2546} \\

\bottomrule
\end{tabular}
\label{tab:composite_editing}}
\end{table}

\subsection{Longform Editing}
Following prior work on long-form model editing, we evaluate on the \textit{WikiBio} and \textit{UnKE} datasets under a long-form generation setting. Unlike short-form editing, which only evaluates whether edited facts appear in short continuations (i.e., next-token or limited-length generation), long-form editing requires the model to consistently maintain and integrate factual updates throughout extended natural language generation, such as full biographies \cite{rosati-etal-2024-long}. For instance, given the query \texttt{``Eleanor Arnason is an American science fiction and fantasy writer''} and an edited knowledge is a long statement such as \texttt{``She is best known for her novel A Woman of the Iron People (1991), which won the James Tiptree, Jr. Award and was a finalist for the Nebula Award for Best Novel.''} The performance of LGA-selected layers on the \textit{UnKE} dataset is presented in Table \ref{tab:unstructured_editing} while \textit{WikiBio} dataset is presented in Figure~\ref{fig:lga-cma} and Table~\ref{tab:combined_tables_editing_methods}, where LGA overall outperforms CMA and SaLEM.

\subsection{Unstructured Editing}\label{app:unstructured_editing}
We perform unstructured model editing on the \textit{UnKE} dataset, adhering to recent protocols and reporting BLEU \cite{papineni2002bleu} (Bilingual Evaluation Understudy) and MMLU (general knowledge accuracy benchmark score) along with various rogue scores \cite{lin2004rouge} like: ROUGE-1 (unigram overlap score), ROUGE-2 (bigram overlap score), ROUGE-L (longest common subsequence-based score), Para-ROUGE-L (ROUGE-L score of paraphrased queries), BERTScore (embedding-based semantic similarity score) as evaluation metrics \cite{deng2025everything}. Table~\ref{tab:unstructured_editing} presents the performance of layers selected by LGA achieve the highest overall score on GPT2-XL and Llama2-7B, while remaining competitive on Gemma3-12B, outperforming CMA and SaLEM in 2 out of 3 settings. Layer selection is carried out using a proxy set of size 15, while evaluation is performed on a test set of size 135 randomly sampled from the \textit{UnKE} dataset. For the task, the Tukey constant is set to 0.5.

\begin{table}[htbp]
\centering
\small
\setlength{\tabcolsep}{2.5pt}
\caption{Unstructured editing performance of LGA-selected layers, SaLEM, and CMA on the \textit{UnKE} dataset across three models: {GPT-2 XL}, {LLaMA2-7B}, and {Gemma3-12B}. We use R-ROME for editing and evaluate using BLEU, ROUGE-1 (R-1), ROUGE-2 (R-2), ROUGE-L (R-L), Para-ROUGE-L (Para-R-L), BERTScore, and MMLU metrics. Layer selection uses a proxy set of size 15, and evaluation is performed on 135 randomly sampled test examples from the \textit{UnKE} dataset. Layers selected by LGA best overall score on GPT2-XL and Llama2-7B, and remain competitive on Gemma3-12B, winning 2 out of 3 settings compared to CMA and SaLEM.}
\resizebox{0.92\textwidth}{!}{
\begin{tabular}{c | c | c c c c c c c c}
\toprule
Model & Method & BLEU $\uparrow$ & R-1 $\uparrow$ & R-2 $\uparrow$ & R-L $\uparrow$ & Para-R-L $\uparrow$ & BertScore $\uparrow$ & MMLU $\uparrow$ & Overall $\uparrow$ \\
\midrule

\multirow{3}{*}{\shortstack[c]{GPT2-XL}}
& CMA    & 0.3182 & 0.1940 & 0.0732 & 0.1841 & 0.1678 & 0.6125 & 0.2133 & 0.2519 \\
& SaLEM  & 0.3474 & 0.2219 & 0.0990 & 0.2097 & 0.1685 & 0.6926 & 0.2207 & 0.2800 \\
& LGA    & 0.3452 & 0.2272 & 0.1036 & 0.2179 & 0.1855 & 0.6895 & 0.2237 & \textbf{0.2846} \\
\midrule
\multirow{3}{*}{\shortstack[c]{Llama2-7B}}
& CMA    & 0.5735 & 0.5310 & 0.3889 & 0.5152 & 0.3776 & 0.8250 & 0.4178 & 0.5184 \\
& SaLEM  & 0.5435 & 0.4837 & 0.3228 & 0.4654 & 0.3610 & 0.8005 & 0.4015 & 0.4826 \\
& LGA    & 0.6342 & 0.6492 & 0.5508 & 0.6398 & 0.3230 & 0.8594 & 0.4296 & \textbf{0.5837} \\
\midrule
\multirow{3}{*}{\shortstack[c]{Gemma3-12B}}
& CMA    & 0.4060 & 0.2971 & 0.1309 & 0.2788 & 0.2229 & 0.6774 & 0.6741 & 0.3839 \\
& SaLEM  & 0.4836 & 0.4105 & 0.2321 & 0.3862 & 0.3084 & 0.7541 & 0.7289 & \textbf{0.4720} \\
& LGA    & 0.3856 & 0.2578 & 0.1017 & 0.2394 & 0.2197 & 0.6501 & 0.6963 & 0.3644 \\
\bottomrule
\end{tabular}
\label{tab:unstructured_editing}}
\end{table}

\section{Comparison with Random Layer Selection} \label{app:random_comparison}
Table~\ref{tab:random_selection_comparison} compares the performance of layers selected by LGA with randomly selected layers across three models on the \textit{ZSRE} dataset. For the random baseline, three layers are sampled uniformly at random and the mean performance is reported. As shown in the table, LGA consistently outperforms random selection in overall performance across all models, with notable improvements in rewrite and rephrase accuracy.

\begin{table}[htbp]
\centering
\small
\setlength{\tabcolsep}{2.5pt}
\caption{Editing performance evaluation of LGA-selected layers and randomly selected layers, denoted as \textit{Random}, for the \textit{ZSRE} dataset across three models: {GPT-2 XL}, {LLaMA2-7B}, and {Gemma3-12B} using R-ROME. For the random baseline, three layers are selected uniformly at random, and their mean performance is reported. The table presents performance metrics, namely \textit{Rewrite Accuracy}, \textit{Rephrase Accuracy}, \textit{Locality}, \textit{Portability}, and \textit{Overall}. Layers for LGA are identified using the proxy set, and evaluation is conducted on an unseen test set, demonstrating performance improvements over random selection.}
\resizebox{0.89\textwidth}{!}{
\begin{tabular}{ c | c | c | c c c c c}
\toprule
Dataset & Model & Method & Rewrite $\uparrow$ & Rephrase $\uparrow$ & Locality $\uparrow$ & Portability $\uparrow$ & Overall $\uparrow$ \\
\midrule

\multirow{9}{*}{\shortstack[c]{ZSRE}}
& \multirow{3}{*}{\shortstack[c]{GPT-2 XL}}
& CMA & 0.9665 & 0.7467 & 0.9608 & 0.4837 & 0.6942 \\
& & Random & 0.8539  & 0.5916 &  0.9701 &     0.4784 &  0.6347 \\
&      & LGA & {0.9851} & {0.8151} & 0.9543 & {0.4757}  & \textbf{0.7106} \\
\cmidrule(lr){2-8}
& \multirow{3}{*}{LLAMA2-7B}
& CMA & 0.9587 & 0.8859 & {0.9923} & 0.5827  & 0.8549 \\
& & Random &0.9576  & 0.8115&   0.9885    &  0.5650    &  0.8307 \\
&      & LGA & {0.9625} & {0.8988} & {0.9927} & {0.5950}  & \textbf{0.8623} \\
\cmidrule(lr){2-8}
& \multirow{3}{*}{GEMMA3-12B}
& CMA & 0.8581 & 0.7041 & {0.9421} & 0.4989 & 0.7508 \\
& & Random&0.8197  & 0.7692 &  0.7988    &  0.4465     & 0.7086 \\
&      & LGA  & {0.9535} & {0.8796} & 0.7736 & {0.5198} & \textbf{0.7816} \\

\bottomrule
\end{tabular}
\label{tab:random_selection_comparison}}
\end{table}

\section{LGA Without Cross-Knowledge Interaction} \label{app:l2norm_comparison}
SaLEM considers only the gradient of the new knowledge $K'_i$. To ensure a fair comparison using only the information utilized by SaLEM, we further evaluate a simplified variant of LGA that removes cross-knowledge interaction between the old-knowledge context $(Q_i \cup K_i)$ and the new-knowledge context $(Q_i \cup K'_i)$, and instead selects layers based solely on the $\ell_2$-norm of the gradient computed from $(Q_i \cup K'_i)$, denoting as LGA($K'$). From Table \ref{tab:l2norm_comparison}, compared to SaLEM, the $\ell_2$-norm variant achieves a small relative improvement of approximately $+1.6\%$ in overall performance, while LGA yields a substantial improvement of approximately $+13.3\%$. These results further highlight the importance of modeling cross-knowledge interactions between old and new knowledge for effective layer selection.

\begin{table}[htbp]
\centering
\small
\setlength{\tabcolsep}{2.5pt}
\caption{Editing performance comparison between CMA, SaLEM, a simplified $\ell_2$-norm variant of LGA, and the full LGA method for layer selection on the \textit{WikiBio} dataset across three models: {GPT-2 XL}, {LLaMA2-7B}, and {Gemma3-12B} using R-ROME. The $\ell_2$-norm variant removes cross-knowledge interaction between the old-knowledge context $(Q_i \cup K_i)$ and the new-knowledge context $(Q_i \cup K'_i)$, and instead selects layers based solely on the gradient computed from $(Q_i \cup K'_i)$, denoting as LGA($K'$). The table reports \textit{Rewrite Accuracy}, \textit{Rephrase Accuracy}, \textit{Locality}, \textit{Portability}, and \textit{Overall editing performance}. Results show that while the $\ell_2$-norm variant achieves a small improvement over SaLEM, LGA significantly outperforms both methods, highlighting the importance of modeling cross-knowledge interactions for effective layer selection.}
\resizebox{0.72\textwidth}{!}{%
\begin{tabular}{ c | c | c | c c c c}
\toprule
Dataset & Model & Method & Rewrite $\uparrow$ & Locality $\uparrow$ & Fluency $\uparrow$ & Overall $\uparrow$ \\
\midrule

\multirow{12}{*}{\shortstack[c]{WikiBio}}
& \multirow{4}{*}{GPT-2 XL}
& CMA & 0.6894  & 0.5692  & 0.8927 & 0.7171 \\
&      & SaLEM & {0.4061}  & 0.2563  & 0.8186 & 0.4937 \\
&      & LGA ($K'$) & {0.4061}  & 0.2563  & 0.8186 & 0.4937 \\
&      & LGA & 0.7484  & 0.5587  & 0.8906 & {0.7326} \\
\cmidrule(lr){2-7}
& \multirow{4}{*}{LLaMA2-7B}
& CMA & 0.8818  & 0.6515  & 0.8547 & 0.7960 \\
& & SaLEM & 0.8396&   0.6526&  0.8529&  0.7817\\
&      & LGA ($K'$) & 0.9560  & 0.6357  & 0.8597 & {0.8172} \\
&      & LGA & 0.9560  & 0.6357  & 0.8597 & {0.8172} \\
\cmidrule(lr){2-7}
& \multirow{4}{*}{GEMMA3-12B}
& CMA & 0.6817  & 0.6546  & 0.6683 & 0.6682 \\
& & SaLEM &0.8153&	0.8062	&	0.6668&	{0.7628}\\
&      & LGA ($K'$) & 0.8124  & 0.8001  & 0.6666 & {0.7597} \\
&      & LGA & 0.8124  & 0.8001  & 0.6666 & {0.7597} \\

\bottomrule
\end{tabular}}
\label{tab:l2norm_comparison}
\end{table}

\section{Leave-One-Dataset-Out Proxy Selection}\label{app:lodo}
For every test set across the datasets (\textit{ZSRE}, \textit{WikiBio}, \textit{WikiCounterfact}, \textit{WikiRecent}, and \textit{Counterfact}) and LLMs (GPT-2 XL, LLaMA2-7B, Gemma3-12B), we evaluate the \textit{Rewrite Accuracy} of the editing layers selected by the (independent) proxy set of each of the other datasets. We then compare this independent proxy-selected layer performance to the layer selected by the (dependent) proxy set drawn from the same dataset as the test set. Note that the performance in both cases is evaluated on the same fixed target test set. Table~\ref{tab:lodo} reports the results.

Across these settings, we observe that the independent proxy set layer performance remains very close to the dependent proxy set layer performance, with an average gap of 0.0179. This indicates cross-dataset generalizability of the proxy set based on \textit{Rewrite Accuracy}.

\begin{table}[htbp]
\centering
\small
\setlength{\tabcolsep}{3pt}
\caption{Cross-dataset generalizability of the proxy set selected golden layer via leave-one-dataset-out evaluation with GPT-2 XL, LLaMA2-7B, and Gemma3-12B on the \textit{ZSRE}, \textit{WikiBio}, \textit{WikiCounterfact}, \textit{WikiRecent}, and \textit{Counterfact} test sets. For each evaluation test set, we report the \textit{Rewrite Accuracy} of the layers selected by the (independent) proxy sets drawn from the other datasets, their mean, and that of the layer selected by the (dependent) proxy set drawn from the same dataset as the test set. The independent proxy sets attain performance close to the dependent proxy set.}
\begin{tabular}{c c c c c c c}
\toprule
\multicolumn{7}{c}{Evaluation Test set: \textit{ZSRE}} \\
\midrule
& \multicolumn{5}{c}{Independent proxy} & Dependent proxy \\
\cmidrule(lr){2-6}\cmidrule(lr){7-7}
Model & \textit{WikiBio} & \textit{WikiCounterfact} & \textit{WikiRecent} & \textit{Counterfact} & Mean & \textit{ZSRE} \\
\midrule
GPT-2 XL & 0.9993 & 0.9935 & 0.9989 & 0.9976 & 0.9973 & 0.9962 \\
LLaMA2-7B & 0.9621 & 0.9650 & 0.9650 & 0.9587 & 0.9627 & 0.9654 \\
Gemma3-12B & 0.9754 & 0.9754 & 0.9317 & 0.9488 & 0.9578 & 0.9328 \\
\midrule
\midrule
\multicolumn{7}{c}{Evaluation Test set: \textit{WikiBio}} \\
\midrule
& \multicolumn{5}{c}{Independent proxy} & Dependent proxy \\
\cmidrule(lr){2-6}\cmidrule(lr){7-7}
Model & \textit{ZSRE} & \textit{WikiCounterfact} & \textit{WikiRecent} & \textit{Counterfact} & Mean & \textit{WikiBio} \\
\midrule
GPT-2 XL & 0.7813 & 0.7484 & 0.8192 & 0.8019 & 0.7877 & 0.8256 \\
LLaMA2-7B & 0.9838 & 0.9851 & 0.9864 & 0.8396 & 0.9487 & 0.9835 \\
Gemma3-12B & 0.7050 & 0.8522 & 0.8448 & 0.6807 & 0.7707 & 0.8522 \\
\midrule
\midrule
\multicolumn{7}{c}{Evaluation Test set: \textit{WikiCounterfact}} \\
\midrule
& \multicolumn{5}{c}{Independent proxy} & Dependent proxy \\
\cmidrule(lr){2-6}\cmidrule(lr){7-7}
Model & \textit{ZSRE} & \textit{WikiBio} & \textit{WikiRecent} & \textit{Counterfact} & Mean & \textit{WikiCounterfact} \\
\midrule
GPT-2 XL & 0.9537 & 0.9433 & 0.9491 & 0.9498 & 0.9490 & 0.9380 \\
LLaMA2-7B & 0.9907 & 0.9893 & 0.9922 & 0.9806 & 0.9882 & 0.9913 \\
Gemma3-12B & 0.8827 & 0.9340 & 0.9380 & 0.9056 & 0.9151 & 0.9340 \\
\midrule
\midrule
\multicolumn{7}{c}{Evaluation Test set: \textit{WikiRecent}} \\
\midrule
& \multicolumn{5}{c}{Independent proxy} & Dependent proxy \\
\cmidrule(lr){2-6}\cmidrule(lr){7-7}
Model & \textit{ZSRE} & \textit{WikiBio} & \textit{WikiCounterfact} & \textit{Counterfact} & Mean & \textit{WikiRecent} \\
\midrule
GPT-2 XL & 0.9845 & 0.9846 & 0.9789 & 0.9855 & 0.9834 & 0.9887 \\
LLaMA2-7B & 0.9779 & 0.9760 & 0.9780 & 0.9674 & 0.9748 & 0.9783 \\
Gemma3-12B & 0.9232 & 0.9789 & 0.9789 & 0.9316 & 0.9532 & 0.9784 \\
\midrule
\midrule
\multicolumn{7}{c}{Evaluation Test set: \textit{Counterfact}} \\
\midrule
& \multicolumn{5}{c}{Independent proxy} & Dependent proxy \\
\cmidrule(lr){2-6}\cmidrule(lr){7-7}
Model & \textit{ZSRE} & \textit{WikiBio} & \textit{WikiCounterfact} & \textit{WikiRecent} & Mean & \textit{Counterfact} \\
\midrule
GPT-2 XL & 0.9796 & 0.9871 & 0.9689 & 0.9828 & 0.9796 & 0.9796 \\
LLaMA2-7B & 0.9979 & 0.9952 & 0.9984 & 0.9979 & 0.9974 & 0.9969 \\
Gemma3-12B & 0.9930 & 0.9866 & 0.9866 & 0.9952 & 0.9904 & 0.9946 \\
\bottomrule
\end{tabular}
\label{tab:lodo}
\end{table}

We further conduct the same evaluation using the \textit{Overall} metric. Since \textit{Overall} aggregates the metrics supported by each dataset (Appendix~\ref{app:metrics}), we compare only datasets that support the same set of metrics: \textit{ZSRE} and \textit{Counterfact} (\textit{Rewrite Accuracy}, \textit{Rephrase Accuracy}, \textit{Locality}, and \textit{Portability}), and \textit{WikiCounterfact} and \textit{WikiRecent} (\textit{Rewrite Accuracy}, \textit{Locality}, \textit{Portability}, and \textit{Fluency}). As shown in Table~\ref{tab:lodo_overall}, the independent proxy set layer performance remains close to the dependent proxy set layer performance, with an average gap of 0.0197.

\begin{table}[htbp]
\centering
\small
\setlength{\tabcolsep}{4pt}
\caption{Cross-dataset generalizability of the proxy set via leave-one-dataset-out evaluation using the \textit{Overall} metric with GPT-2 XL, LLaMA2-7B, and Gemma3-12B, between datasets that support the same set of metrics. For each evaluation test set, we report the \textit{Overall} performance of the layer selected by the (independent) proxy set drawn from the other dataset and that of the layer selected by the (dependent) proxy set drawn from the same dataset as the test set. The independent proxy set attains performance close to the dependent proxy set.}
\begin{tabular}{c c c c c}
\toprule
& \multicolumn{2}{c}{Evaluation Test set: \textit{ZSRE}} & \multicolumn{2}{c}{Evaluation Test set: \textit{Counterfact}} \\
\cmidrule(lr){2-3}\cmidrule(lr){4-5}
& Independent proxy & Dependent proxy & Independent proxy & Dependent proxy \\
Model & \textit{Counterfact} & \textit{ZSRE} & \textit{ZSRE} & \textit{Counterfact} \\
\midrule
GPT-2 XL & 0.8106 & 0.8286 & 0.7789 & 0.7726 \\
LLaMA2-7B & 0.8623 & 0.8638 & 0.8077 & 0.8255 \\
Gemma3-12B & 0.7564 & 0.8261 & 0.6722 & 0.7361 \\
\midrule
\midrule
& \multicolumn{2}{c}{Evaluation Test set: \textit{WikiCounterfact}} & \multicolumn{2}{c}{Evaluation Test set: \textit{WikiRecent}} \\
\cmidrule(lr){2-3}\cmidrule(lr){4-5}
& Independent proxy & Dependent proxy & Independent proxy & Dependent proxy \\
Model & \textit{WikiRecent} & \textit{WikiCounterfact} & \textit{WikiCounterfact} & \textit{WikiRecent} \\
\midrule
GPT-2 XL & 0.6788 & 0.6916 & 0.7087 & 0.7108 \\
LLaMA2-7B & 0.7609 & 0.7656 & 0.7604 & 0.7625 \\
Gemma3-12B & 0.6209 & 0.6411 & 0.6307 & 0.6474 \\
\bottomrule
\end{tabular}
\label{tab:lodo_overall}
\end{table}

\section{Limitations}\label{app:limitations}
% Limitations is added as per paper's checklist at the end
As we show through extensive experiments across multiple LLM architectures, benchmark datasets, and editing methods, our proposed Layer Gradient Analysis (LGA) approach outperforms CMA in editing performance and computational efficiency. As a result, our work enables more reliable, efficient, and targeted updates to LLMs, which is critical as these models are increasingly deployed in real-world applications. However, as with any work, there are limitations to our approach that can be addressed in future work. For instance, a limitation that can be addressed is the need for further study and benchmarks of golden layer estimation and knowledge editing methods in LLMs, extending beyond the models, datasets, and editing strategies explored in this work. Future work could also explore strategies to make LGA more robust to extreme or unusual gradients, further improving its reliability and applicability across various LLMs and datasets.}

\section{Broader Impacts}\label{app:broader_impacts}
Our work on knowledge editing in LLMs offers a promising direction for enabling efficient and targeted updates to model knowledge. There are many potential positive societal consequences of our work, such as enabling LLMs to efficiently incorporate updated knowledge without requiring brute-force re-training, improved editing performance at scale with lower computational overhead, as well as more robust and reliable models, among other benefits. Since knowledge editing requires access to model internals and can only be undertaken by model developers, there are no specific societal risks associated with our proposed approaches as such. Furthermore, our approach provides insights into which layers are most responsible for incorporating new knowledge, offering a practical perspective on model behavior and novel mechanistic insights. These insights may guide future research toward more effective and efficient knowledge editing in language models.

% Our work primarily focuses on the knowledge editing problem in LLMs, and hence, generally advances the field of machine learning and generative artificial intelligence. There are many potential positive societal consequences of our work, enabling LLMs to efficiently incorporate updated knowledge utilizing current and new knowledge without requiring computationally expensive re-training, improved editing performance and efficiency in identification of effective editing layers at scale with lower computational overhead, as well as more robust and reliable models, among other benefits. In general, none of the potential societal consequences of our work lead to a detrimental impact, and none exist that necessitate a comprehensive discussion here.

% https://anonymous.4open.science/r/Golden-Layers-LGA-6D08/
\section{Code and Reproducibility}\label{app:code}
We provide our code and implementation in an open-source repository: \url{https://github.com/Shrestha30/Golden-Layers-LGA}. All the experiments were conducted on a Linux server with 6x NVIDIA DGX B200 GPUs with 192 GB VRAM/GPU. All the utilized LLMs are original (unquantized) open-source versions from HuggingFace. During generation, we used greedy decoding without sampling. To ensure reproducibility, all sources of randomness, including PyTorch, NumPy, and Python’s random module, were fixed using a seed of 42. The gradient of the loss with respect to the model’s existing knowledge was computed for each input sequence, considering up to the number of tokens in the ground-truth target if available, or up to number of tokens in the new target knowledge. Our codebase was built upon the \textit{EasyEdit} \cite{wang2024easyediteasytouseknowledgeediting} framework, and evaluations of the edited models were performed following their standard conventions.

%%%%%%%%%%%%%%%%%%%%%%%%%%%%%%%%%%%%%%%%%%%%%%%%%%%%%%%%%%%%

\newpage
\section*{NeurIPS Paper Checklist}

\begin{enumerate}

\item {\bf Claims}
    \item[] Question: Do the main claims made in the abstract and introduction accurately reflect the paper's contributions and scope?
    \item[] Answer: \answerYes{}%\answerTODO{} % Replace by \answerYes{}, \answerNo{}, or \answerNA{}.
    \item[] Justification: The abstract and introduction accurately reflect the paper's contributions: empirical demonstration of golden layers, their generalizability via proxy datasets, and the LGA method for efficient golden layer identification. All claims are validated by experiments in Sections 4 and 5.%\justificationTODO{}

\item {\bf Limitations}
    \item[] Question: Does the paper discuss the limitations of the work performed by the authors?
    \item[] Answer: \answerYes{}%\answerTODO{} % Replace by \answerYes{}, \answerNo{}, or \answerNA{}.
    \item[] Justification: Appendix \ref{app:limitations} provides a discussion on Limitations. %\justificationTODO{}

\item {\bf Theory assumptions and proofs}
    \item[] Question: For each theoretical result, does the paper provide the full set of assumptions and a complete (and correct) proof?
    \item[] Answer: \answerNA{}%\answerTODO{} % Replace by \answerYes{}, \answerNo{}, or \answerNA{}.
    \item[] Justification: The paper is empirical in nature and does not include formal theoretical results, theorems, or proofs. The contributions are validated experimentally.%\justificationTODO{}

    \item {\bf Experimental result reproducibility}
    \item[] Question: Does the paper fully disclose all the information needed to reproduce the main experimental results of the paper to the extent that it affects the main claims and/or conclusions of the paper (regardless of whether the code and data are provided or not)?
    \item[] Answer: \answerYes{}%\answerTODO{} % Replace by \answerYes{}, \answerNo{}, or \answerNA{}.
    \item[] Justification: Appendices \ref{app:code} and \ref{app:implementation} describe relevant details, e.g. hardware, libraries, random seeds, decoding strategy, implementation details, etc. enabling full reproducibility. %\justificationTODO{}

\item {\bf Open access to data and code}
    \item[] Question: Does the paper provide open access to the data and code, with sufficient instructions to faithfully reproduce the main experimental results, as described in supplemental material?
    \item[] Answer: \answerYes{}%\answerTODO{} % Replace by \answerYes{}, \answerNo{}, or \answerNA{}.
    \item[] Justification: An anonymized open-source repository is provided at (\url{https://anonymous.4open.science/r/Golden-Layers-LGA-6D08/}). All datasets used are publicly available benchmarks (ZSRE, WikiBio, WikiCounterfact, WikiRecent, Counterfact).%\justificationTODO{}

\item {\bf Experimental setting/details}
    \item[] Question: Does the paper specify all the training and test details (e.g., data splits, hyperparameters, how they were chosen, type of optimizer) necessary to understand the results?
    \item[] Answer: \answerYes{}%\answerTODO{} % Replace by \answerYes{}, \answerNo{}, or \answerNA{}.
    \item[] Justification: Appendices \ref{app:code} and \ref{app:implementation} provide implementation details. %\justificationTODO{}

\item {\bf Experiment statistical significance}
    \item[] Question: Does the paper report error bars suitably and correctly defined or other appropriate information about the statistical significance of the experiments?
    \item[] Answer: \answerYes{}%\answerTODO{} % Replace by \answerYes{}, \answerNo{}, or \answerNA{}.
    \item[] Justification: Table \ref{tab:oracle_vs_golden_datasets} reports mean ± standard deviation for golden layer vs. sample-wise optimal layer comparisons, with two-sided Student's t-test results explicitly reported. Figure \ref{fig:proxy_vs_test} includes the standard error of the mean as error bars.%\justificationTODO{}

\item {\bf Experiments compute resources}
    \item[] Question: For each experiment, does the paper provide sufficient information on the computer resources (type of compute workers, memory, time of execution) needed to reproduce the experiments?
    \item[] Answer: \answerYes{}%\answerTODO{} % Replace by \answerYes{}, \answerNo{}, or \answerNA{}.
    \item[] Justification: Appendix \ref{app:code} specifies the compute environment: a Linux server with 6× NVIDIA DGX B200 GPUs, each with 192 GB VRAM. The paper also reports runtime comparisons between methods in Figure \ref{fig:runtime_analysis}.%\justificationTODO{}
    
\item {\bf Code of ethics}
    \item[] Question: Does the research conducted in the paper conform, in every respect, with the NeurIPS Code of Ethics \url{https://neurips.cc/public/EthicsGuidelines}?
    \item[] Answer: \answerYes{}%\answerTODO{} % Replace by \answerYes{}, \answerNo{}, or \answerNA{}.
    \item[] Justification: Code of ethics has been reviewed. %\justificationTODO{}

\item {\bf Broader impacts}
    \item[] Question: Does the paper discuss both potential positive societal impacts and negative societal impacts of the work performed?
    \item[] Answer: \answerYes{}%\answerTODO{} % Replace by \answerYes{}, \answerNo{}, or \answerNA{}.
    \item[] Justification: Yes, broader impacts are discussed in Appendix \ref{app:broader_impacts}.%\justificationTODO{}
    
\item {\bf Safeguards}
    \item[] Question: Does the paper describe safeguards that have been put in place for responsible release of data or models that have a high risk for misuse (e.g., pre-trained language models, image generators, or scraped datasets)?
    \item[] Answer: \answerNA{}%\answerTODO{} % Replace by \answerYes{}, \answerNo{}, or \answerNA{}.
    \item[] Justification: The paper proposes a layer selection method for knowledge editing and does not require specific safeguards.%\justificationTODO{}

\item {\bf Licenses for existing assets}
    \item[] Question: Are the creators or original owners of assets (e.g., code, data, models), used in the paper, properly credited and are the license and terms of use explicitly mentioned and properly respected?
    \item[] Answer: \answerYes{}%\answerTODO{} % Replace by \answerYes{}, \answerNo{}, or \answerNA{}.
    \item[] Justification: All datasets and models are cited with original references. %\justificationTODO{}

\item {\bf New assets}
    \item[] Question: Are new assets introduced in the paper well documented and is the documentation provided alongside the assets?
    \item[] Answer: \answerNo{}%\answerTODO{} % Replace by \answerYes{}, \answerNo{}, or \answerNA{}.
    \item[] Justification: No new assets released.%\justificationTODO{}

\item {\bf Crowdsourcing and research with human subjects}
    \item[] Question: For crowdsourcing experiments and research with human subjects, does the paper include the full text of instructions given to participants and screenshots, if applicable, as well as details about compensation (if any)? 
    \item[] Answer: \answerNA%\answerTODO{} % Replace by \answerYes{}, \answerNo{}, or \answerNA{}.
    \item[] Justification: The paper involves no crowdsourcing and no research with human subjects. All experiments are conducted programmatically on LLMs using benchmark datasets.%\justificationTODO{}

\item {\bf Institutional review board (IRB) approvals or equivalent for research with human subjects}
    \item[] Question: Does the paper describe potential risks incurred by study participants, whether such risks were disclosed to the subjects, and whether Institutional Review Board (IRB) approvals (or an equivalent approval/review based on the requirements of your country or institution) were obtained?
    \item[] Answer: \answerNA{}%\answerTODO{} % Replace by \answerYes{}, \answerNo{}, or \answerNA{}.
    \item[] Justification: No human subjects research was conducted. IRB approval is not required for this work.%\justificationTODO{}

\item {\bf Declaration of LLM usage}
    \item[] Question: Does the paper describe the usage of LLMs if it is an important, original, or non-standard component of the core methods in this research? Note that if the LLM is used only for writing, editing, or formatting purposes and does \emph{not} impact the core methodology, scientific rigor, or originality of the research, declaration is not required.
    %this research? 
    \item[] Answer: \answerNA{}%\answerTODO{} % Replace by \answerYes{}, \answerNo{}, or \answerNA{}.
    \item[] Justification: This research does not involve LLMs as any important, original, or non-standard components.%\justificationTODO{}

\end{enumerate}

\end{document}